\documentclass[twoside]{article}

\usepackage[accepted]{aistats2023}
\usepackage[authoryear]{natbib}

\usepackage[autonum,colorhypersetup]{shortex}

\usepackage{amsfonts}
\usepackage{float}
\usepackage{graphicx}
\graphicspath{ {./figures/} }
\usepackage{caption}
\usepackage{cleveref}
\usepackage{lipsum}

\newtheorem{condition}{Condition}

\usepackage{xcolor}
\usepackage{algorithm} 
\usepackage{algpseudocode}
\usepackage{float}
\usepackage{enumitem}
\usepackage{etoolbox}
\usepackage{authblk}

\newcommand{\wheretext}[1]{\textcolor{magenta}{[#1]}}

\newcommand{\diagnosticName}{TADDAA\xspace}

\newcommand{\posterior}{\Pi}
\newcommand{\posteriorDensity}{\pi}

\newcommand{\initialDensity}{\hat{\pi}^{(0)}}
\newcommand{\improvedDensityAt}[1]{\hat{\pi}^{(#1)}}
\newcommand{\improvedDensity}{\improvedDensityAt{T}}

\newcommand{\proposal}[3]{Q_{#3}(#1, \ifstrempty{#2}{\cdot}{\dee{#2}})}  %
\newcommand{\proposalDensity}[3]{q_{#3}(#1, #2)}  %
\newcommand{\mhkernel}[3]{P_{#3}(#1, \ifstrempty{#2}{\cdot}{\dee{#2}})}  %
\newcommand{\proposalParam}{\psi}
\newcommand{\acceptprob}[2]{\alpha(#1, #2)}
\newcommand{\rejectprob}[2]{r_{#2}(#1)}

\newcommand{\state}{x}
\newcommand{\proposedState}{y}
\newcommand{\stateDim}{d}

\newcommand{\mcProposedState}{Y}

\newcommand{\CurrentSamples}{X_{1:N}^{(t)}}
\newcommand{\CurrentProposals}{Y_{1:N}^{(t)}}

\newcommand{\CurrentSingleSample}{X_{j}^{(t)}}
\newcommand{\CurrentSingleSamplePre}{X_{j}^{(t-1)}}
\newcommand{\CurrentSingleProposal}{Y_{j}^{(t)}}
\newcommand{\CurrentSingleProposalPre}{Y_{j}^{(t-1)}}

\newcommand{\CurrentStepsize}{h_{N}^{(t)}}
\newcommand{\PreviousStepsize}{h_{N}^{(t-1)}}

\newcommand{\LimitingStepsizePre}{\bar{h}^{(t-1)}}
\newcommand{\LimitingStepsize}{\bar{h}^{(t)}}

\newcommand{\EmpiricalMeasureX}{\nu_{N}^{(t)}}
\newcommand{\LimitingMeasureXt}{\bar{\nu}^{(t)}}

\newcommand{\EmpiricalMeasureXY}{\xi_{N}^{(t)}}
\newcommand{\EmpiricalMeasureXYPre}{\xi_{N}^{(t-1)}}
\newcommand{\LimitingMeasureXY}{\bar{\xi}^{(t)}}
\newcommand{\LimitingMeasureXYPre}{\bar{\xi}^{(t-1)}}

\newcommand{\EmpiricalMeasureXYTilde}{\tilde{\xi}_N^{(t)}}

\newcommand{\Filtrationcurrent}{\mathcal{F}_{N}^{(t)}}
\newcommand{\FiltrationcurrentPre}{\mathcal{F}_{N}^{(t-1)}}

\begin{document}

\twocolumn[

\aistatstitle{A Targeted Accuracy Diagnostic for Variational Approximations}

\aistatsauthor{ Yu Wang \And  Miko\l{}aj Kasprzak \And  Jonathan H. Huggins}

\aistatsaddress{ Boston University \And  MIT and University of Luxembourg \And Boston University} 

]

\begin{abstract}
	Variational Inference (VI) is an attractive alternative to Markov Chain Monte Carlo (MCMC) due 
	to its computational efficiency in the case of large datasets and/or complex models with high-dimensional parameters. 
	However, evaluating the accuracy of variational approximations remains a challenge.
	Existing methods characterize the quality of the whole
	variational distribution, which is almost always poor in realistic applications, 
	even if specific posterior functionals such as the component-wise means or variances are accurate.
	Hence, these diagnostics are of practical value only in limited circumstances. 
	To address this issue, we propose the \emph{TArgeted Diagnostic for Distribution Approximation Accuracy} (\diagnosticName),
	which uses many short parallel MCMC chains to obtain lower bounds on the error of each posterior functional of interest.
	We also develop a reliability check for \diagnosticName to determine when the lower bounds should not be trusted. 
	Numerical experiments validate the practical utility and computational efficiency of our approach on a range of synthetic distributions and real-data examples,
	including sparse logistic regression and Bayesian neural network models. 
\end{abstract}

\section{INTRODUCTION}

Bayesian inference is widely used due to its flexibility to handle arbitrary models, the desirable decision-theoretic properties of the provided point estimates, and its coherent
uncertainty quantification \citep{robert2007bayesian}. 
However, posterior point estimates and uncertainties usually cannot be calculated analytically, so in practice users of Bayesian methods rely on approximation algorithms. 
Variational inference (VI) has become an increasingly popular alternative to Markov chain Monte Carlo (MCMC) 
due to its computational efficiency and the availability of black-box versions that can be applied to nearly arbitrary models \citep{jordan1999introduction, wainwright2008graphical, hoffman2013stochastic, kingma2013auto, blei2017variational,  kucukelbir2017automatic, agrawal2020advances}. 

The computational efficiency of variational inference arises from optimizing a divergence between the exact posterior and a distribution constrained to a tractable approximating family. 
The trade-off is that even the optimal distribution within this family may be a poor approximation to the posterior, so in general there is no guarantee that variational point estimates and uncertainties will be accurate.
Therefore, it is important to develop diagnostic methods that quantify the error introduced by using a variational approximation. 
However, existing diagnostic methods 
either lack interpretability \citep{gorham2017measuring,mackey2016measuring, wada2015stopping, chee2017convergence, burda2015importance, domke2018importance}, or
do not provide useful information when the parameter space is high-dimensional \citep{yao2018yes,xing2020distortion,huggins2020validated}.
For example, \citet{yao2018yes} propose using Pareto smoothed importance sampling (PSIS) $\hat{k}$, which essentially quantifies how effective the approximation is as a proposal distribution for importance sampling -- not whether it will provide good approximations to specific quantities of interest. 
The kernel Stein discrepancy \citep[KSD;][]{gorham2017measuring, mackey2016measuring} can be used to evaluate the discrepancy between the approximate and the true distribution but 
its value is not readily interpretable and depends on a distant dissipativity condition that is not always easy to verify in practice. 
VI quality can also be accessed by monitoring the running average of changes in the estimated objective, which is usually the evidence lower bound (ELBO) \citep{wada2015stopping, chee2017convergence, burda2015importance, domke2018importance}.
However, like the KSD, the ELBO lacks interpretability because it includes an unknown constant offset (the marginal likelihood), which is hard to estimate well
unless the approximation is very accurate.
Moreover, all these methods attempt to characterize the accuracy of the full variational distribution, which in practice is usually poor even if specific summaries 
such as posterior means, variances, or credible intervals are accurate. 
Therefore, in practice, they provide limited useful diagnostic information in the types of complex models where variational methods offer the greatest computational savings. 

In this paper, we propose a diagnostic that is applicable to high-dimensional parameter spaces and can provide lower bounds on the error of specific posterior summaries, including marginal means and covariances. 
Our approach is to run many short MCMC chains with stationary distribution equal to the posterior $\posteriorDensity$.
These chains are initialized with independent samples from a posterior approximation $\initialDensity$, which, after $T$ iterations, results 
in an improved empirical distribution $\improvedDensity$ that is closer to $\posteriorDensity$. 
If $\initialDensity$ is a poor approximation to $\posteriorDensity$, then the improved distribution $\improvedDensity$ obtained using MCMC can be significantly different from $\initialDensity$;
otherwise, the distribution will not shift.
Thus, we can obtain lower bounds on the error of a posterior summary by computing confidence intervals for how much that summary differs under $\initialDensity$ and $\improvedDensity$. 
Our approach might fail to provide meaningful results, however, if the Markov kernel used mixes poorly. 
To ameliorate this issue, we jointly adapt the Markov chains at each iteration.
We show that if the number of chains is large, they remain essentially independent despite the joint adaptation procedure; hence, our confidence intervals remain valid. 
We also propose a simple correlation-based check that can detect poorly mixing chains, and hence warn the user against trusting the diagnostic. 
Numerical experiments validate the practical utility and computational efficiency of our approach on a range of synthetic distributions and real-data examples.\footnote{A \texttt{Python}  implementation of \diagnosticName and code to reproduce all of our experiments is available at \url{https://github.com/TARPS-group/TADDAA}.}

\section{PRELIMINARIES}
\label{sec:background}

Modern Bayesian statistics relies heavily on models with difficult-to-compute posterior distributions, and hence requires 
general-purpose algorithms to approximate expectations with respect to those distributions. 
Throughout we abuse notation and use $\posteriorDensity$ to denote the posterior distribution and the corresponding density. 

\paragraph{Markov Chain Monte Carlo.}
Markov chain Monte Carlo (MCMC) sampling methods provide a general-purpose framework for obtaining samples that are asymptotically exact (in runtime).
To construct a Markov kernel with the desired stationary distribution, a standard approach is to correct an arbitrary
proposal kernel using the Metropolis--Hastings (MH) accept-reject procedure \citep{metropolis1953equation}. 
For $\state \in \reals^{\stateDim}$, let $\proposal{\state}{\proposedState}{\proposalParam}$ denote the proposal kernel parameterized by $\proposalParam$ 
with current state $\state$ and corresponding density $\proposalDensity{\state}{\proposedState}{\proposalParam}$. 
A proposed state $\mcProposedState \distas \proposal{\state}{}{\proposalParam}$ is accepted with probability 
\[ \label{eq:MH-accept}
\acceptprob{\state}{\mcProposedState} 
= \min \left\{1, \frac{\posteriorDensity(\mcProposedState) \proposalDensity{\mcProposedState}{\state}{\proposalParam}}{\posteriorDensity(\state) \proposalDensity{\state}{\mcProposedState}{\proposalParam}}\right\}. 
\]
The Metropolis--Hastings Markov transition kernel is therefore given by 
\[ \label{eq:mh-kernel}
\mhkernel{\state}{\proposedState}{\proposalParam} 
&\defas \acceptprob{\state}{\proposedState} \proposal{\state}{\proposedState}{\proposalParam}+ \rejectprob{\state}{\proposalParam} \delta_{\state}(\dee\proposedState),
\]
where $\rejectprob{\state}{\proposalParam} \defas 1 - \int \acceptprob{\state}{\proposedState} \proposal{\state}{\proposedState}{\proposalParam}$ is the rejection probability 
and $\delta_{\state}$ is a Dirac measure at $\state$. 

\paragraph{Variational Inference.}
Variational inference (VI) provides a potentially faster alternative to MCMC when models are complex and/or the dataset size is large \citep{blei2017variational}.
Variational inference aims to minimize some measure of discrepancy $\mathcal{D}_{\posteriorDensity}(\cdot)$ over a tractable family $\mathcal{Q}$ of potential
approximating distributions \citep{wainwright2008graphical}, resulting in a posterior approximation given by 
\[
\hat{\posteriorDensity}   = \argmin_{\xi \in \mathcal{Q}} \mathcal{D}_{\posteriorDensity}(\xi).
\]
We will focus on the \emph{Kullback--Leibler (KL) divergence}
\[
\mathcal{D}_{\posteriorDensity}(\xi) = \mathrm{KL}(\xi \mid \posteriorDensity) :=\int \log \left(\frac{\mathrm{d} \xi}{\mathrm{d} \posteriorDensity}\right) \mathrm{d} \xi
\]
as the discrepancy since it is the most widely used: 
the unknown marginal likelihood does not affect the optimization, computing gradients requires estimating expectations only with respect to distribution within a tractable family $\mathcal{Q}$,
and optimization is relatively stable \citep{blei2017variational}. %

\paragraph{Alternative Non-exact Approximation Methods.}
While our experiments focus on variational inference, numerous other non-exact posterior approximation methods are widely used in practice
such as expectation-propagation \citep{Minka:2001,bishop2006pattern}, the Laplace approximation \citep{gelman1995bayesian,bishop2006pattern}, and 
the iterated nested Laplace approximation \citep{Rue:2009,GomezRubio:2020:INLA}. 
Our diagnostic applies equally well to all of these methods. 

\section{AN MCMC-BASED ACCURACY DIAGNOSTIC}
\label{MCMC diagnostics}

\begin{figure}[t]
	\centering
	\includegraphics[width=\linewidth,clip,trim=1in 2.5in 0.6in 0.92in]{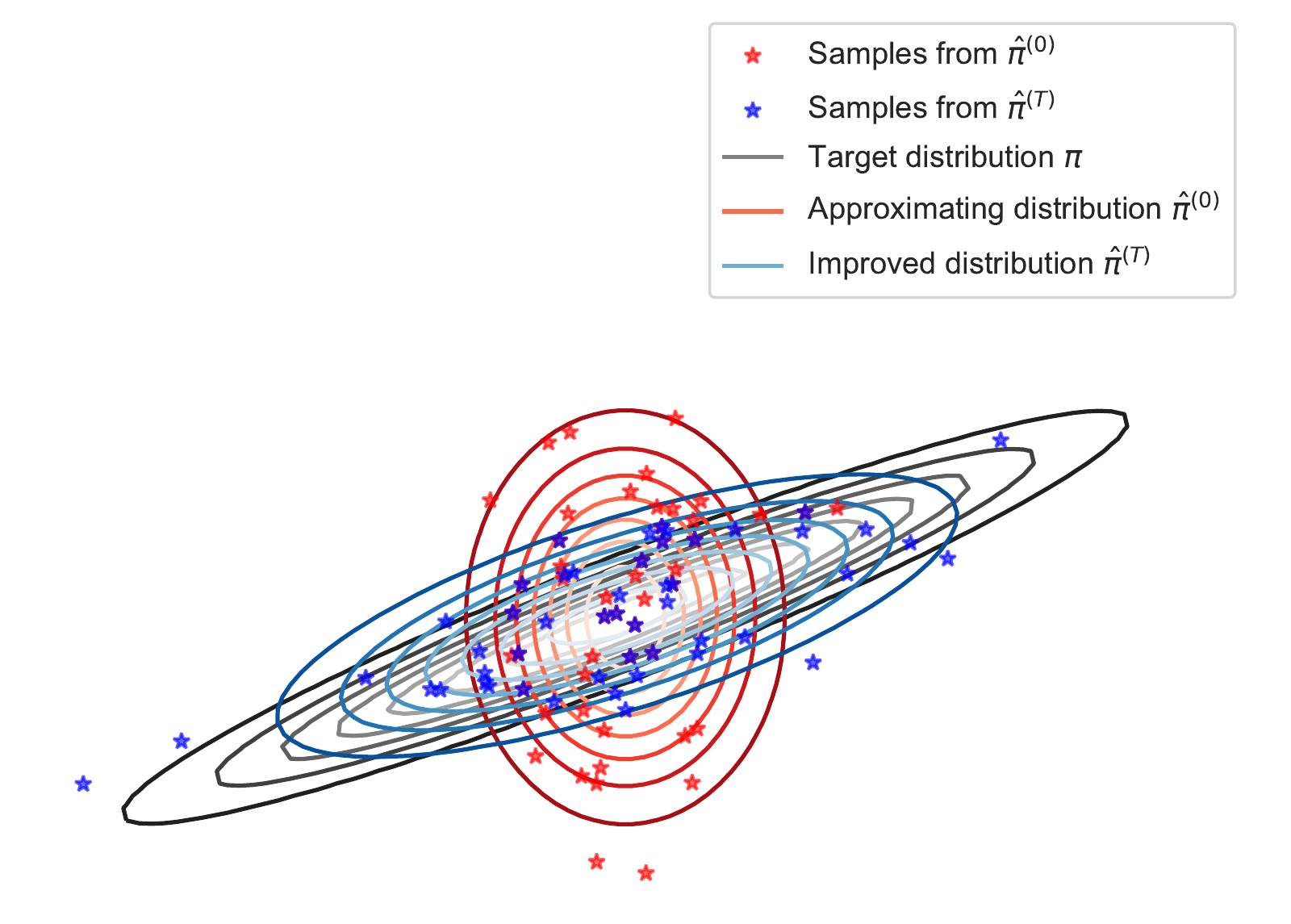}
	\caption{A cartoon plot for MCMC-based accuracy diagnostic. 
		Even if the Markov chains do not reach the stationary distribution, they can provide a lower bound on error in the posterior approximation.}
	\label{fig:cartoon}
\end{figure}

Suppose we have an approximation $\initialDensity$ to the posterior density $\posteriorDensity$. 
Given a sample $X^{(0)} \distas \initialDensity$, we could improve it by applying a Markov kernel $\mhkernel{\state}{\proposedState}{\proposalParam}$ with invariant distribution $\posteriorDensity$, resulting in a sample $X^{(1)} \distas \mhkernel{X^{(0)}}{}{\proposalParam}$ with a distribution $\improvedDensityAt{1}$ that is typically closer to $\posteriorDensity$ than
$\initialDensity$ if $\initialDensity \ne \posteriorDensity$. 
Continuing to run the Markov chain for $T$ total steps results in a sample $X^{(T)}$ with distribution $\improvedDensity$  that, when $\initialDensity$ is a poor approximation to $\posteriorDensity$,
can be substantially closer to $\posteriorDensity$; see \cref{fig:cartoon} for an illustration.  
More specifically, consider some posterior functional of interest such as a mean or variance, which we generically denote by $\mcF$.
For example, $\mcF(\posteriorDensity) = \int \state_{i} \posteriorDensity(\dee\state)$ would be the mean of the $i$th component. 
If the initial approximation error $\varepsilon^{(0)}(\mcF) \defas \mcF(\initialDensity) - \mcF(\posteriorDensity)$ is large, 
then a well-chosen Markov kernel will result in the improved approximation error $\varepsilon^{(T)}(\mcF) \defas \mcF(\improvedDensity) - \mcF(\posteriorDensity)$
being substantially smaller.
In this case, $|\mcF(\initialDensity) - \mcF(\improvedDensity)|$ measures the decrease in the initial approximation error, and hence provides a lower bound on $|\varepsilon^{(0)}(\mcF)|$. 
If, on the other hand, the initial approximation error is small, then $|\mcF(\initialDensity) - \mcF(\improvedDensity)|$ should remain small as well. 

While in principle the just-described approach is appealing, there are two immediate problems.
First, we do not have direct access to $\improvedDensity$.
Second, $\mhkernel{\state}{\proposedState}{\proposalParam}$ may not be particularly efficient unless the proposal parameter $\proposalParam$ is well-chosen --
which is nearly impossible to do \emph{a priori}.
To address both of these problems, we propose to run $N$ Markov chains in parallel and, following standard MCMC practice, adapt the proposal parameter at each iteration \citep{andrieu2008tutorial}. 
Denote the state of the $j$th chain after iteration $t$ by $X_{j}^{(t)}$ and the proposal parameter used at iteration $t$ by $\proposalParam^{(t-1)}$. 
We write $X_{1:N}^{(t)} = (X_{j}^{(t)})_{j=1}^{N}$ and similarly for $X_{j}^{(1:t)}$, $X_{1:N}^{(1:t)}$, and other quantities. 
In the case of a Metropolis--Hastings kernel, at iteration $t$, for each $j \in \{1,\dots,N\}$, we generate proposals 
$Y_{j}^{(t+1)} \distas \proposal{X_{j}^{(t)}}{}{\proposalParam^{(t)}}$, then accept or reject the proposal to obtain $X_{j}^{(t+1)}$.  
Next, we can use \emph{inter-chain adaptation} \citep[INCA;][]{craiu2009learn}, which uses all the samples available up to the current time to update the proposal parameter
to $\proposalParam^{(t+1)} = H(\proposalParam^{(1:t)}, X_{1:N}^{(1:t+1)}, Y_{1:N}^{(1:t+1)})$.
We call $H$ the \emph{adaptation function}. 

Finally, we can use the $N$ samples from iteration $T$ to construct a $(1-\alpha)$-confidence interval $(\ell_{\mcF}, u_{\mcF})$ for $\mcF(\initialDensity) - \mcF(\improvedDensity)$, 
and hence a high-confidence lower bound on $|\mcF(\initialDensity) - \mcF(\posteriorDensity)|$,
if the following condition holds:
\begin{condition} \label{cond:monotonicity}
	The map $t \mapsto |\varepsilon^{(t)}(\mcF)|$ is non-increasing and the sign of $\varepsilon^{(t)}(\mcF)$ does not depend on $t$. 
\end{condition}
Under \cref{cond:monotonicity}, with probability $1-\alpha$, 
\[
\begin{aligned}
	\label{eq:general lower bound}
	\varepsilon^{(0)}(\mcF)
	&\ge |\mcF(\initialDensity) - \mcF(\improvedDensity)| \\
	&\ge \begin{cases}
		0 & \text{if $\ell_{\mcF} \le 0 \le u_{\mcF}$} \\
		\min(|\ell_{\mcF}|, |u_{\mcF}|) & \text{otherwise} 
	\end{cases} \\
	&= \ind\left\{0 \notin (\ell_{\mcF},  u_{\mcF})\right\} \times \min(|\ell_{\mcF}|, |u_{\mcF}|) \\
	&\defines  B_{\mcF}. 
\end{aligned}
\]
\cref{alg:diagnostic-skeleton} summarizes the key steps of our proposed \emph{TArgeted Diagnostic for Distribution Approximation Accuracy} (\diagnosticName) 
for computing the error lower bound $B_{\mcF}$. 
References to specific implementation recommendations (all described and justified in detail below) are in magenta. 

\begin{remark}
	While in general we expect \cref{cond:monotonicity} to hold, it is possible that it could be violated for target distributions with complex geometries.
	Although we only consider reversible Markov kernels, it is plausible that non-reversible Markov kernels could also lead to violations. 
\end{remark}

\begin{algorithm}[t]
	\caption{\diagnosticName}
	\label{alg:diagnostic-skeleton}
	\textbf{Input:} $\log$ density of the target $\posteriorDensity$, \\
	\phantom{\textbf{Input:}\,} approximating distribution $\initialDensity$, \\
	\phantom{\textbf{Input:}\,} functional of interest $\mcF$, \\
	\phantom{\textbf{Input:}\,} proposal kernel $\proposal{\state}{\proposedState}{\proposalParam}$ \wheretext{\cref{sec:kernel-choice}}, \\
	\phantom{\textbf{Input:}\,} initial proposal parameter $\proposalParam^{(0)}$  \wheretext{\cref{sec:kernel-choice}},  \\
	\phantom{\textbf{Input:}\,} number of Markov chains $N$ \wheretext{\cref{eq:sample-size}}, \\
	\phantom{\textbf{Input:}\,} number of iterations $T$ \wheretext{\cref{eq:chain-length} or \eqref{eq:chain-length-HMC}}
	\begin{algorithmic}[1]
		\State $X_{j}^{(0)} \distas \initialDensity$ for $j = 1,\dots,N$
		\For{\texttt{$t=0$ to $T-1$}}
		\For{\texttt{$j=1$ to $N$}}
		\State $Y_{j}^{(t+1)} \distas \proposal{X_{j}^{(t)}}{}{\proposalParam^{(t)}}$
		\State $\alpha^{(t)}_{j} = \acceptprob{X_{j}^{(t)}}{Y_{j}^{(t)}}$ 
		\State $X_{j}^{(t+1)} = \begin{cases}
			Y_{j}^{(t)}, & \text{with probability } \alpha_{j}^{(t)} \\ 
			X_{j}^{(t)}, & \text{with probability } 1-\alpha_{j}^{(t)}
		\end{cases}$
		\EndFor
		\State $\proposalParam^{(t+1)} = H(\proposalParam^{(1:t)},X_{1:N}^{(1:t+1)}, Y_{1:N}^{(1:t+1)})$ \wheretext{\cref{eq:adaptive step size}}
		\EndFor
		\State Use $\initialDensity$ (or $X_{1:N}^{(1)}$) and $X_{1:N}^{(T)}$ to compute a $(1-\alpha)$-confidence interval 
		$(\ell_{\mcF}, u_{\mcF})$ for $\mcF(\initialDensity) - \mcF(\improvedDensity)$
		\State Compute correlation check $\rho^{2}_{\max}(T)$ to verify reliability of the diagnostic \wheretext{\cref{sec:reliability-check}}
		\State \Return  $B_{\mcF}$ as defined in \cref{eq:general lower bound} and $\rho^{2}_{\max}(T)$
	\end{algorithmic}
\end{algorithm}

We will construct confidence intervals under the following condition.
\begin{condition}
	\label{cond:CLT}
	The elements of $X_{1:N}^{(T)}$ are identically distributed, have finite second moments, and are pairwise 
	independent. 
\end{condition}
We will focus on diagnosing the accuracy of means, variances, and quantiles.
But \diagnosticName can be straightforwardly applied to other functionals of interest. 
For $t \in \{0,\dots, T\}$, define the mean vector
$\mu^{(t)} = (\mu_{1}^{(t)}, \ldots, \mu_{d}^{(t)})^{\top} \defas  \EE_{X \sim \improvedDensityAt{t}}(X)$, 
and marginal standard deviations $\sigma^{(t)}_{i} \defas \operatorname{stdev}_{X \sim \improvedDensityAt{t}}(X_{i})$ for $i = 1,\dots,d$. %
\paragraph{Marginal mean.}
For the mean functional $\mcF_{\text{mean}}^{i}(\posteriorDensity) \defas \mu_{i}(\posteriorDensity) \defas \int x_{i}\,\posteriorDensity(\dee x)$, 
under \cref{cond:CLT}, a $(1-\alpha)$ confidence interval for $\mu_{i}^{(T)}-\mu_{i}^{(0)}$ is given by \citep{mendenhall2012introduction}
\[
\label{eq: one sample mean difference ci}
\overline{x}_{i}^{(T)}-\mu^{(0)}_{i} \pm \frac{s_{i}^{(T)}}{\sqrt{N}} t_{N-1}(1-\alpha / 2),
\]
where %
$\overline{x}_{i}^{(T)}= \frac{1}{N} \sum_{j=1}^{N} x_{j, i}^{(T)}$ is the sample mean, 
$
s_{i}^{(T)}= \frac{1}{N-1}\big\{\sum_{j=1}^{N}(x_{j, i}^{(T)}-\overline{x}_{i}^{(T)})^{2}\big\}^{1/2}
$
is the sample standard deviation, and $t_{N-1}(q)$ is the $q$ quantile of a $t$ distribution with $N -1$ degrees of freedom. 
\paragraph{Marginal variance.}
For the variance functional $\mcF_{\text{var}}^{i}(\posteriorDensity) \defas \sigma_{i}^{2}\defas \var_{X \distas \posteriorDensity}(X_{i})$,
under \cref{cond:CLT}, a $(1-\alpha)$ confidence interval for $2\log ( \sigma_{i}^{(T)}/ \sigma_{i}^{(0)})$ is given by \citep{mendenhall2012introduction}
\[
\label{one sample var ratio ci}
\textstyle   \left[ \log \left( \frac{(N-1)s_{i}^{(T)}}{( \sigma_{i}^{(0)})^{2}\chi^{2}_{N-1}(1-\alpha/2)} \right),  \log \left( \frac{(N-1)s_{i}^{(T)}}{( \sigma_{i}^{(0)})^{2}\chi^{2}_{N-1}(\alpha/2)} \right) \right], 
\]
where
$\chi^{2}_{N-1}(q)$ is the $q$ quantile of a $\chi^{2}$ distribution with $N -1$ degrees of freedom. 

\paragraph{Marginal quantile.}
	For the $p$-quantile functional 
	\[
	\label{eq: quantile functionals of interest}
	\!\!\!\!\mcF^{i}_{p}(\pi) = Q^{i}_{p}(\pi) \defas \inf \{q \in \reals: P_{X \sim \pi}(X_{i}\leq q) \geq p\}, \!\!
	\]
	under \cref{cond:CLT}, a $(1-\alpha)$ confidence interval for $Q^{i}_{p}(\improvedDensity)-Q^{i}_{p}(\initialDensity)$ is given by  \citep{hahn2011statistical}
	\[
	\left[X^{(T)}_{(l), i}-Q^{i}_{p}(\initialDensity), X^{(T)}_{(u), i}-Q^{i}_{p}(\initialDensity)\right],
	\]
	where $X^{(T)}_{(1), i}\leq X^{(T)}_{(2), i} \leq \cdots \leq X^{(T)}_{(N), i}$ are the order statistics of the samples $X_{1:N}^{(T)}$, $l=B(\alpha/2, N, p)$, $u=B(1-\alpha/2, N, p)+1$,  and $B(q, N, p)$ is the $q$ quantile of a binomial distribution with parameters $N$ and $p$.

While we have now described the key components of \diagnosticName, a practical and reliable implementation requires addressing a number of issues: 
\begin{enumerate}
	\item \textbf{Proposal kernel and adaptation scheme.} 
	The choice of MCMC proposal kernel and adaptation function $H$ can have a dramatic effect on sampling efficiency and thus the accuracy of \diagnosticName. 
	We propose to use gradient-based proposals that scale well to high-dimensional problems and focus on adapting the step size parameter of these algorithms. 
	(\cref{sec:kernel-choice})
	\item \textbf{Inter-chain dependence.} INCA introduces dependence between the Markov chains, which could invalidate the confidence intervals. 
	However, we show that the samples from the Markov chains at any fixed iteration satisfy a \emph{propagation-of-chaos} property, which
	implies that the chains are asymptotically independent (in the limit $N \to \infty$), so an asymptotic version of \cref{cond:CLT} still holds. (\cref{sec:poc-clt})
	\item \textbf{Choice of $T$ and $N$.} The {length of each Markov chain $T$} and the {number of Markov chains $N$} must be 
	chosen such that the diagnostic is computationally efficient while 
	still (i) having $T$ large enough that the improved approximation $\improvedDensity$ becomes sufficiently different from the initial approximation $\initialDensity$
	and (ii) having $N$ large enough that the confidence intervals are sufficiently small to provide meaningful diagnostic information. (\cref{sec:MC-size-parameters}) 
	\item \textbf{Checking the reliability of the diagnostic.} Since it is possible the proposal kernel will be inefficient despite adaptation,
	we propose a correlation-based reliability check for our diagnostic. (\cref{sec:reliability-check})
\end{enumerate}

\subsection{Markov Kernels and Adaptation} \label{sec:kernel-choice}

We consider four possible Markov kernels: random walk Metropolis--Hastings (RWMH), Metropolis-adjusted Langevin algorithm (MALA), the Barker algorithm (Barker),
and Hamiltonian Monte Carlo (HMC), which we describe in detail in \cref{sec:MH-kernels}. 
All four kernels rely on a step-size parameter $h$. %
The latter three algorithms exploit gradient information about $\log \posteriorDensity$, and thus have superior efficiency when $d$ is large.
High-dimensional sampling efficiency can be formalized using the theory of \emph{optimal scaling} \citep{Roberts:2001}. 
To guarantee sampling efficiency and obtain well-behaved limiting behavior, optimal scaling results show that step size $h$ needs to scale as $\Theta(d^{-\gamma})$, which
results in a mixing time (i.e., number of iterations to obtain an effectively independent sample) of $\Theta(d^{\gamma})$. 
Hence, the smaller $\gamma$ is the more efficient the kernel. 
The values of $\gamma$ for the kernels we consider are $1$ for RWMH \citep{gelman1997weak}, $1/3$ for MALA \citep{roberts1998optimal}, $1/3$ for Barker \citep{livingstone2019barker}, and $1/4$ for HMC \citep{beskos2013optimal}. 
Optimal scaling theory also provides the optimal asymptotic acceptance rates ($\bar{\alpha}_{*}$) for each kernel, which are 0.234 for RWMH, 0.4 for Barker, 0.574 for MALA, and 0.651 for HMC.
While HMC has the best scaling with $\gamma = 1/4$, the efficiency of Barker is more robust to the precise step size (and hence precise acceptance rate) \citep{livingstone2019barker}, 
which makes it attractive for our use-case where it may not be possible to obtain the optimal step size since adaptation occurs over a relatively small number of iterations $T$. 

We adapt the step size so that the acceptance rate approaches the optimal asymptotic acceptance rate $\bar{\alpha}_{*}$ \citep{andrieu2008tutorial}. 
Let $\bar\alpha^{(t)} \defas N^{-1}\sum_{j=1}^{N}\alpha^{(t)}_{j}$ denote the average acceptance probability at iteration $t$. 
Letting $\psi^{(t)} =\log h^{(t)}$ be the log of the step size, the adaptation function is given by 
\[
\label{eq:adaptive step size}
H(\proposalParam^{(1:t)}, X_{1:N}^{(1:t+1)}, &\,Y_{1:N}^{(1:t+1)}) \\
&= \psi^{(t)} +  \frac{1}{\sqrt{t+1}}(\bar\alpha^{(t)} - \bar{\alpha}_{*}). 
\]
Compared with running many parallel chains independent of each other, 
using a joint adaptation scheme improves adaption speed and ensures the samples at each iteration follow the same distribution \citep{rosenthal2000parallel,solonen2012efficient}.
The sequence of proposal parameters $(\proposalParam^{(t)})_{t=0}^{\infty}$ converge under the standard Robbins--Monro conditions (see \citet[Section 4]{andrieu2008tutorial} for a related discussion). In order to prove convergence to the stationary distribution of the resulting Markov chain, one needs to verify an additional assumption, such as uniform ergodicity \citet[Theorem 1]{roberts2007coupling} or one of the weaker sufficient conditions proposed, for instance in \citet[Assumption 3.1]{atchade2005adaptive} or \citet[Section 2.4]{green2015}. 

Following \citet{roberts1998optimal}, we initialize the step sizes to $h^{(0)} = 2.4^2/d$ for RWMH, and $h^{(0)} = {2.4^2}/{d^{1/3}}$ for MALA and Barker,
and $h^{(0)} = {2.4^2}/{d^{1/4}}$ for HMC.
All of the Markov kernels we consider can make use of a pre-conditioning matrix, which is beneficial when components have different scales \citep{roy2022convergence, stramer2007bayesian}. 
While it is possible to adapt the pre-conditioning matrix in addition to the step size in practice we found it sufficed to keep the pre-conditioning matrix fixed and equal to the 
covariance of $\initialDensity$. 
If the covariance is not available in closed form we use the sample covariance of the samples $X_{1:N}^{(0)}$. 

\subsection{Asymptotic Independence of Adapted Markov Chains}  \label{sec:poc-clt}

Due to the use of adaptation, the Markov chains $\CurrentSamples$ are not independent once $t > 1$, 
so the final samples $X_{1:N}^{(T)}$ violate the independence requirement of \cref{cond:CLT}. 
However, we can instead ensure that the Markov chains are \emph{asymptotically} independent in the following sense:
\begin{definition}
	Let $X_{N,1:N} = (X_{N,1}, \dots, X_{N,N})$ denote a random vector.
	The sequence of random vectors $\{ X_{N,1:N} \}_{N=1}^{\infty}$ is \emph{$\bar\nu$-chaotic} if, 
	for any $r \in \nats$ and any bounded continuous real-valued functions $g_{1}, g_{2}, \ldots, g_{r}$,
	\[
	\begin{aligned}
		\lim _{N \rightarrow \infty}\EE_{X_{N,1:N}}\left\{\prod_{i=1}^{r}g_{i}\left(X_{N,i}\right)\right\}
		&= \prod_{i=1}^{r} \int g_{i}(x) \bar\nu(\dee x).
	\end{aligned}
	\]
\label{chaos definition}
\end{definition}

In particular, we show that the Markov chains are chaotic after any fixed number of iterations. 
\begin{theorem}
	\label{propagation of chaos for x}
	Under Assumption \ref{assumption}  (given below), for any $t \in \nats$, there exists a probability 
	distribution $\LimitingMeasureXt$ such that the sequence 
	$\{\CurrentSamples\}_{N=1}^{\infty}$ is $\LimitingMeasureXt$-chaotic.
\end{theorem}
The proof can be found in \cref{proofs}.
Crucially, the asymptotics we require for \cref{propagation of chaos for x} are in the number of Markov chains $N$, \emph{not} the number of iterations $T$. 
Thus, \emph{for any fixed $T$}, \cref{cond:CLT} holds asymptotically in $N$.

To write our assumptions, note that 
if we treat the proposed states $\CurrentProposals$ as additional random variables, we can write the MCMC update as 
\[
(X_{j}^{(t+1)}, Y_{j}^{(t+1)}) \distas T(X_{j}^{(t)}, Y_{j}^{(t)}, \CurrentStepsize, \cdot, \cdot)
\]
for an appropriate Markov kernel $T(x, y, h, \dee x', \dee y')$, where $h$ is the step size. 
\begin{assumption} 
	\label{assumption}
	\begin{enumerate}[label={\textbf{(\alph*)}}, ref={\theassumption.(\alph*)}]
		\item \label{assumption1}%
		The proposal probability density $q_{h}(y, x)$ is continuous with respect to $(x, y, h)$.
		\item \label{assumption2}%
		The target distribution has a continuous probability density function $\posteriorDensity(\cdot)$.
		\item \label{assumption3}%
		Samples generated from the Markov transition kernel $T (x, y, h, \cdot, \cdot )$ satisfy $\EE \|X_{j}^{(t)}\|^{2} < \infty$ and $\EE \|Y_{j}^{(t)}\|^{2} < \infty$ for any $t \in \nats$.
	\end{enumerate}
\end{assumption}
These conditions are all quite mild.
All of the proposals used in our experiments satisfy Assumption \ref{assumption1}.
Assumption \ref{assumption2} is required to use all three gradient-based proposals (MALA, Barker, and HMC). 
Assumption \ref{assumption3} just requires that for any fixed time $t \in \nats$, the generated samples have finite second moment.
\begin{remark}
	\cref{propagation of chaos for x} could be generalized to targets with discrete components if the proposal distribution is absolutely continuous with respect to an appropriate reference measure 
	and the topology of the space on which the target is defined respects the discrete and continuous components of the target. 
	(For example, if the target distribution is defined on $\reals$ and has an atom at zero, then $\{0\}$ should be an open set.) In that case, \cref{chaos definition} would need to be adapted in order to account for the change in the underlying topology, and so will the details of the proof of \cref{propagation of chaos for x}.
\end{remark}

\subsection{Length and Number of Markov Chains} \label{sec:MC-size-parameters}

Having established the validity of \cref{alg:diagnostic-skeleton}, we now turn to the question of how 
to ensure the computational cost is not prohibitive and the diagnostic is accurate. 
Given a target distribution and Markov kernel, the computational cost is determined by the number of Markov chains $N$ and iterations $T$. 
We address each in turn.

\paragraph{Number of Markov Chains.}
The number of Markov chains must be sufficiently large that the confidence intervals are small enough to detect meaningful errors.
Our approach is to choose $N$ to be the smallest value that satisfies the user's tolerance level for the margin of error.
In the case of mean estimation, it follows from \cref{eq: one sample mean difference ci} that the error for estimating $\mu_{i}^{(T)}$ is given by
$
t_{N-1}(\alpha / 2)\,{s}_{i}^{(T)} /\sqrt{N}.
$
Similarly, the log variance error can also be determined by \cref{one sample var ratio ci}.
For the mean we use the {relative} error normalized by the standard deviation, as this accounts for the relevant scale of the error. 
Hence, if the user's tolerance is $\delta_{\text{mean}}$ for relative mean error and $\delta_{\text{var}}$ for log variance error\footnote{The width of the quantile confidence interval is sample dependent, so we do not set a tolerance level for the quantile error.}, 
then the required sample size (number of Markov chains) is %
\[
\begin{aligned}
	\label{eq:sample-size}
	\textstyle  N &= \textstyle  \max \left( N_{\text{mean}}, N_{\text{variance}}\right),
\end{aligned}
\]
where 
\[
\textstyle N_{\text{mean}} &\defas \textstyle  \min \left\{ n \in \nats : \frac{t_{n-1}(\alpha / 2)}{\sqrt{n}} \leq \delta_{\text{mean}} \right\},  \\
\textstyle  N_{\text{variance}} &\defas  \textstyle  \min \left\{ n \in \nats :  \log\left( \frac{\chi^{2}_{n-1}(1-\alpha/2)}{\chi^{2}_{n-1}(\alpha/2)}\right) \leq \delta_{\text{var}} \right\}.
\]

\paragraph{Number of Iterations.} 
The number of iterations has to be large enough that the improved distribution $\improvedDensity$ is sufficiently different from a poor approximating distribution $\initialDensity$.
As discussed earlier, the theory of optimal scaling shows that the Markov chain requires $\Theta(d^{\gamma})$ iterations to mix.
Thus, we propose to scale $T$ with respect to dimension $d$ in the same manner. 
For RWMH\footnote{Based on optimal scaling behavior of RWMH, \cref{eq:chain-length} may not be large enough to guarantee significant change in the distribution. So, we do not recommend RWMH.}, MALA, and Barker, we take
\[
\label{eq:chain-length}
T = \floor[1]{c d^{1/3}},
\]
while for HMC we take 
\[
\label{eq:chain-length-HMC}
T = \floor[1]{c d^{1/4}/L},
\]
where $L$ is the number of leapfrog steps in HMC. 
Scaling by $L$ is necessary for HMC to ensure $T$ accounts for the $L$ gradient evaluations required per iteration.
All other methods require 0 or 1 gradient evaluation per iteration. 
Based on numerical results we suggest $c = 50$. 
In \cref{sec: ablation study}, we also run ablation studies on how the diagnostic result (several types of lower bound) would change with $T$, which shows that \cref{eq:chain-length} and \cref{eq:chain-length-HMC} are reasonable.

The results of \citet{bhatia2022statistical} suggest our choice of $T$ will ensure the cost of \diagnosticName will not be too large compared to the cost of
using black-box variational inference algorithms. 
Specifically, \citet[Theorem 1]{bhatia2022statistical} shows that the number of gradient evaluations of stochastic optimization needed to obtain a constant optimization error scales as $d^{1/3}$. 
Hence, the computational cost (as measured by number of gradient evaluations) is $\Theta(d^{1/3})$ for VI, $\Theta(d^{1/3})$ for MALA and Barker, and $\Theta(d^{1/4})$ for HMC. 
We verify the cost of \diagnosticName is not prohibitive in practice in our numerical experiments. 
\begin{remark}
	\label{remark: validity of TADDAA}
	\diagnosticName is essentially running MCMC \emph{only} for a burn-in period.
	Crucially, the diagnostic remains valid and useful even if the chains remain far from stationarity
	as long as $\improvedDensity$ is closer to stationarity (i.e., it improves the approximating distribution $\initialDensity$ somewhat). 
	In challenging scenarios where moving closer to stationarity is hard even with best practices (e.g., INCA and preconditioning), 
	the \diagnosticName error lower bounds can be close to zero, leading to uninformative diagnostic results.
	See \cref{sec: experiment of candy classification} for a concrete example.
	We note also that there is a large literature studying the convergence rates of MCMC algorithms \citep{brooks2011handbook, roberts2004general}.
\end{remark}

\subsection{A Reliability Check for the Diagnostic} \label{sec:reliability-check}
The reliability of \diagnosticName depends on the mixing behavior of the Markov chains: 
\begin{itemize}
	\item If the Markov chains are mixing well, we expect the diagnostic results can be trusted.
	\item If the Markov chains are not mixing well, diagnosis of a poor approximation can still be trusted but diagnosis of a good approximation may not be reliable. 
	In the latter case, we should consider increasing the length of Markov chains or otherwise improving the Markov kernel. %
\end{itemize}
The usual approach to checking MCMC mixing is some version of the $\widehat{R}$ diagnostic \citep{gelman1992inference,vats2021revisiting, vehtari2021rank}. 
However, using $\widehat{R}$ requires a sufficiently large effective sample size for each Markov chain, which is not satisfied for the short Markov chains used in \diagnosticName. 
Instead, we propose to leverage the fact that we have many nearly independent chains to compute the correlation between the initial and final values of each chain.
If the chains are mixing, then this correlation will be close to zero. 
Specifically we propose to use the worst-case correlation coefficient
$
\rho^{2}_{\max}(T)
\defas \max_{i} \rho^{2}_{i}(T),
$
where 
\[
\begin{aligned}
	\textstyle \rho^{2}_{i}(T)
	\defas  \frac{\sum_{j=1}^{N} ( X_{j,i}^{(0)}-\hat{\mu}_{i}^{(0)})( X_{j,i}^{(T)}-\hat{\mu}_{i}^{(T)})}{\sqrt{\sum_{j=1}^{N}( X_{j,i}^{(0)}-\hat{\mu}_{i}^{(0)})^{2}} \sqrt{(\sum_{j=1}^{N} X_{j,i}^{(T)}-\hat{\mu}_{i}^{(T)})^{2}}}.
\end{aligned}
\]
If $\rho^{2}_{\max}(T)$ is close to 0 (we use a cutoff of $0.1$), then the check passes; otherwise, the check fails.
In addition to examining the value of $\rho^{2}_{\max}(T)$, we can also examine plots of $t$ versus $\rho^{2}_{\max}(t)$ and $t$ versus $\rho^{2}_{i}(t)$ to gain insight into,
how quickly mixing is occurring, respectively, overall and for specific coordinates. 

\begin{figure}[t]
	\includegraphics[width=\linewidth]{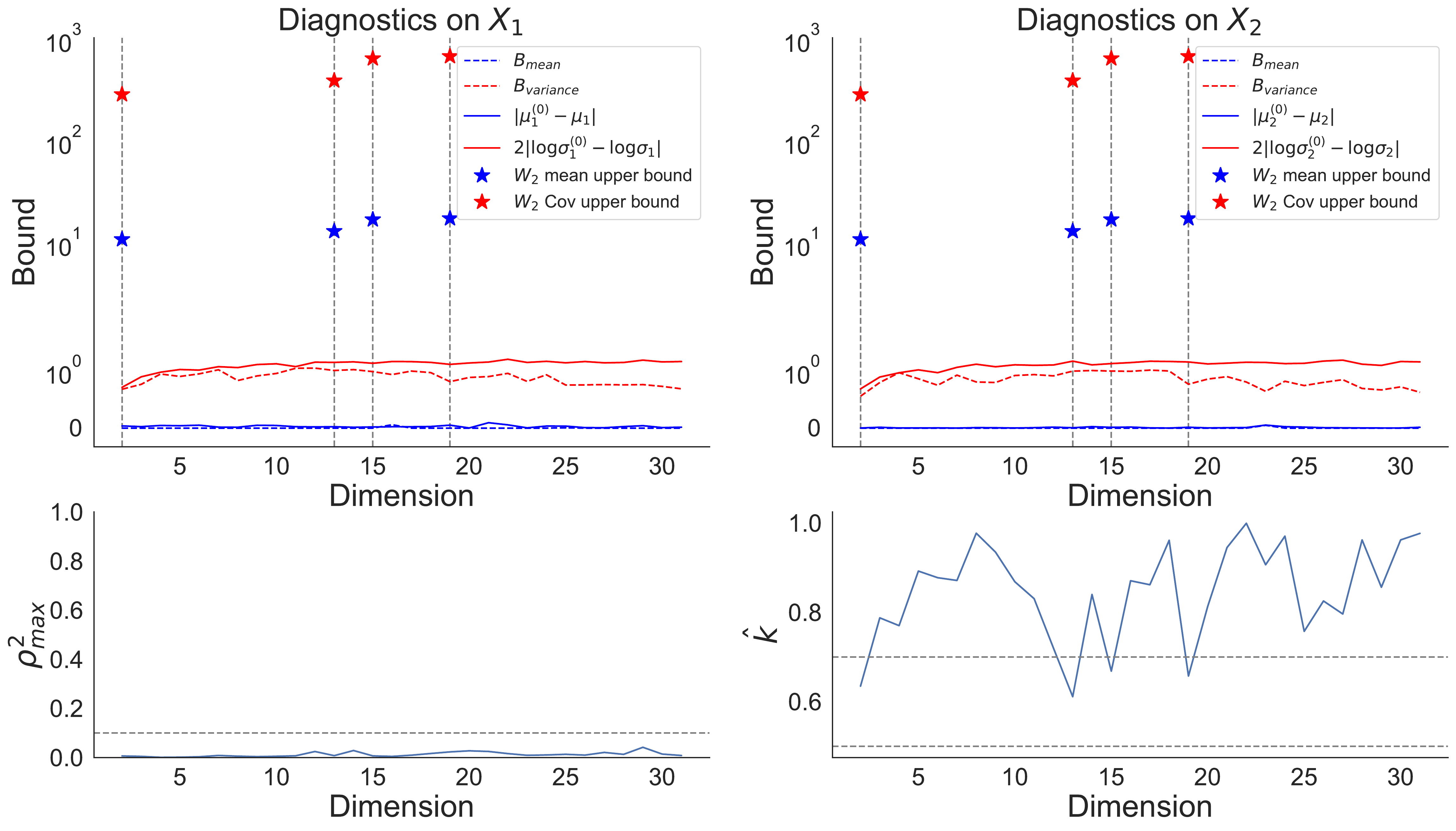}
	\caption{Diagnostics for mean-field VI approximation to correlated Gaussian targets, where
		\diagnosticName uses the Barker proposal. 
		Here $\mu_{i}$ and $\sigma_{i}^{2}$ denote, respectively, the mean and variance of $X_{i}$. }
	\label{fig:High Dimensional Gaussian Diagnostics}
\end{figure}

\begin{figure}[t]
	\includegraphics[width=\linewidth]{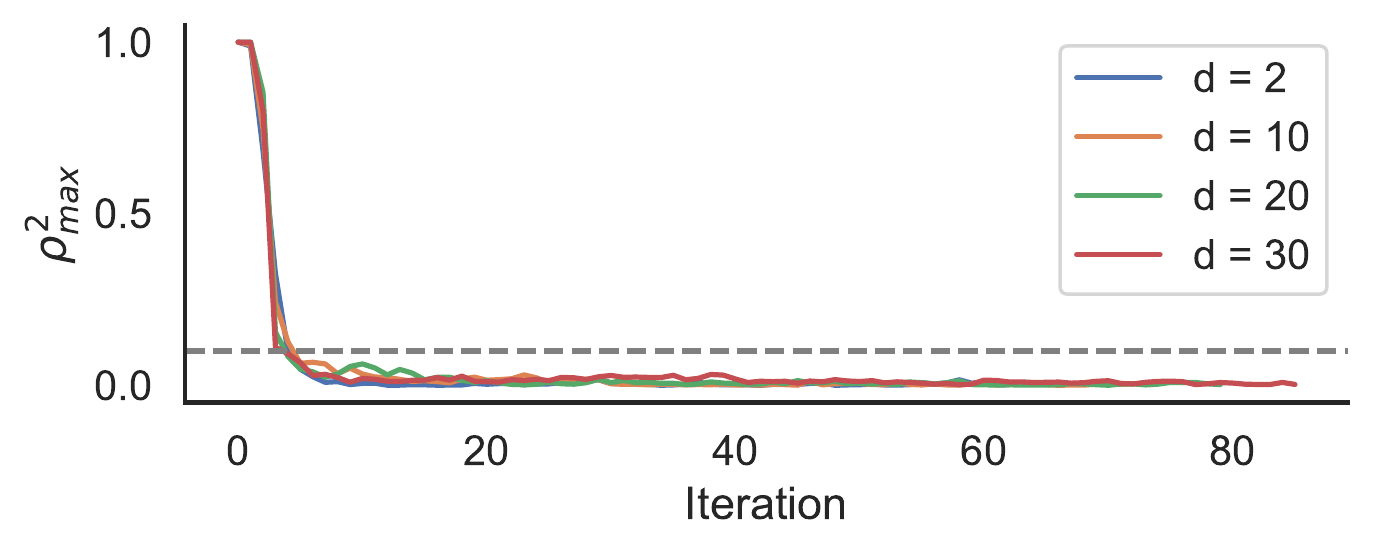}
	\caption{Reliability checks for \diagnosticName using Barker applied to the mean-field VI approximation to correlated Gaussian targets.}
	\label{fig:High Dimensional Gaussian reliability check}
\end{figure}
\section{EXPERIMENTS}
\label{sec: EXPERIMENTS}
We now validate the efficacy and computational efficiency of \diagnosticName and the correlation-based reliability check. 
In our experiments, the number of Markov chains is $N=386$, which corresponds to the tolerance level $\delta_{\text{mean}} = 0.1$ and $\delta_{\text{var}} = 0.15$.
While we often present results for all four kernels discussed in \cref{sec:kernel-choice} (RWMH, MALA, Barker, and HMC),
we generally recommend using Barker because it has the efficiency of gradient-based kernels but is less sensitive to parameter tuning than MALA and HMC. 
For variational inference, we use the default settings of the \textsf{bbvi} function in the VIABEL package \citep{Welandawe:2022:BBVI}, which used a mean-field Gaussian approximation family. 
We set the maximum number of iterations to $200{,}000$
but the optimization can terminate early if convergence is reached. 
We compare \diagnosticName to the $\hat{k}$ diagnostic proposed in \citet{yao2018yes} and the Wasserstein-based mean and variance upper bounds proposed in \citet{huggins2020validated}.

\subsection{Gaussian Model With Correlated Coordinates}
\label{sec: Gaussian Model With Correlated Coordinates}
We first investigate the performance of \diagnosticName in a high-dimensional setting where the variational mean-field approximation has correct
mean but incorrect variance estimates, which is a common situation in practice. 
Specifically, we use a correlated Gaussian target $\posteriorDensity = \distNorm(\mu, \Sigma)$, where $\mu \in \reals^{d}$, 
$\Sigma_{ii} = \sigma_{i}^{2}$, and $\Sigma_{ij} = \rho \sigma_{i}\sigma_{j}$ for $i \ne j$. 
	We introduce correlation by setting $\rho = 0.7$ and variance heterogeneity by setting $\sigma_{1}^{2}=10$ and $\sigma_{i}^{2}=1$ for $i=2,3,...,d$.

	\cref{fig:High Dimensional Gaussian Diagnostics} shows that the lower bounds $B_{\text{mean}}$ and $B_{\text{variance}}$ remain valid and useful even when $d$ and $\hat{k}$ are large: 
	\diagnosticName correctly captures that the mean estimates are accurate but the variance estimates are inaccurate. 
	From \cref{fig:High Dimensional Gaussian reliability check}, we can see that the Barker chains mix well in all cases. %
	\Cref{fig:High Dimensional Gaussian Diagnostics} also provides a comparison to $\hat{k}$ and the Wasserstein-based upper bounds proposed in \citet[Theorem 3.4]{huggins2020validated}.
	The $\hat{k}$ result suggests that the VI approximation is not reliable for almost any dimension (since $\hat{k}>0.7$). 
	But if the functional of interest is the mean, VI should be considered reliable. %
	Similarly, the Wasserstein upper bounds are too conservative by a factor of 10--1,000, incorrectly suggesting the VI approximation is extremely inaccurate.
	Moreover, the upper bounds are unavailable once $\hat{k}>0.7$ \citep{huggins2020validated}.
	Overall, \diagnosticName provides much more precise and actionable information about the approximation quality than the other diagnostics.

	\begin{figure}[t]
		\includegraphics[width=\linewidth]{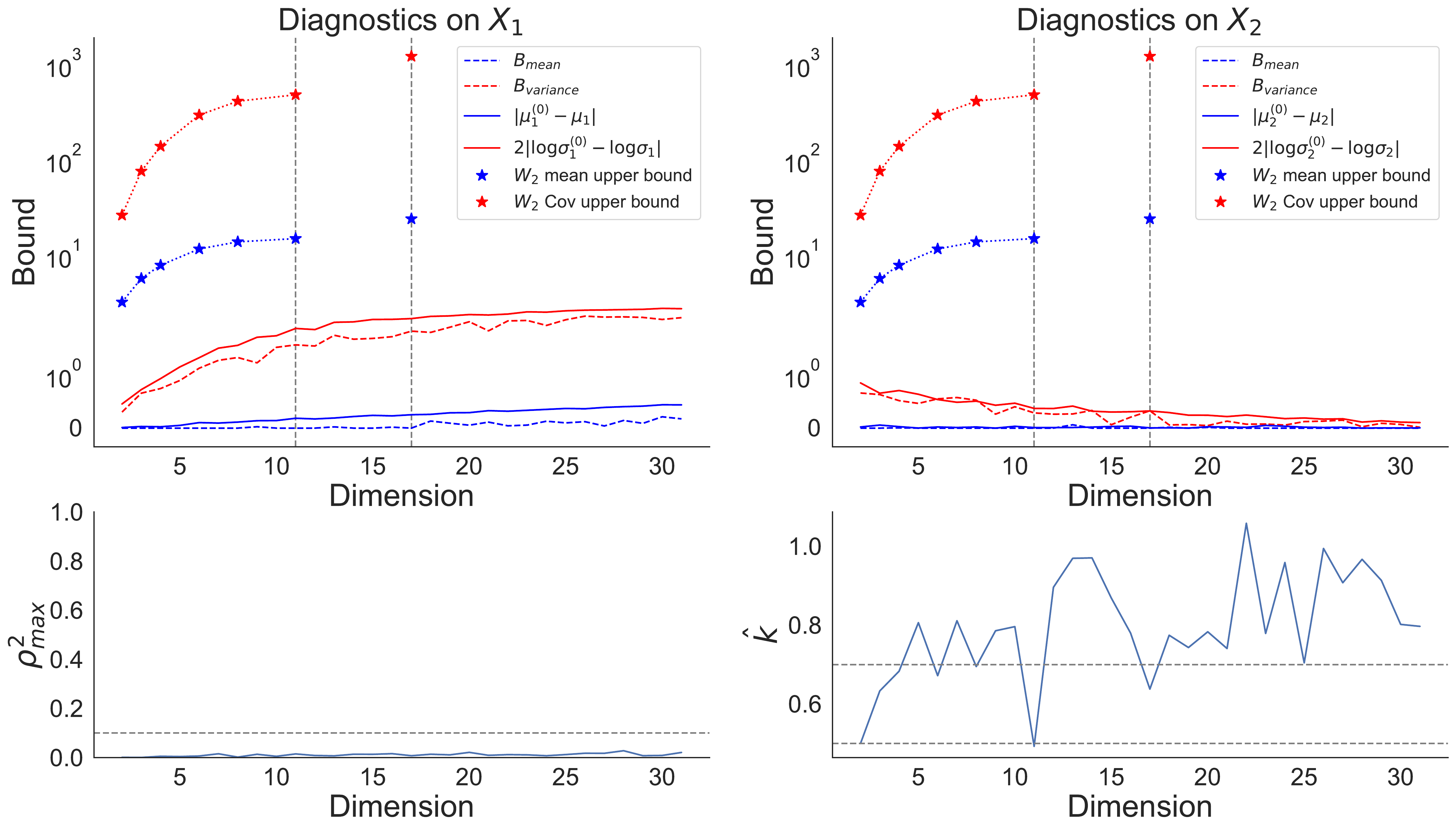}
		\caption{Diagnostics for Neal-funnel shape model, where 
			\diagnosticName uses the Barker proposal. 
			Here $\mu_{i}$ and $\sigma_{i}^{2}$ denote, respectively, the mean and variance of $X_{i}$. }
		\label{fig:High Dimensional Neal Funnel Diagnostics}
	\end{figure}

	\subsection{Neal-Funnel Shape Model}
	\label{sec:Neal-Funnel Shape Model}
	
	Next, we consider a more challenging target with a more complex geometry: Neal's funnel distribution \citep{neal2003slice}, which is similar to the geometries encountered in
	hierarchical models. 
	For $x = (x_{1},\dots,x_{d}) \in \mathbb{R}^{d}$, let $\sigma \defas \exp(x_{1})$. 
	The funnel distribution is given by $\posteriorDensity(x) =  \distNorm(x_{1} \given 0, 1)\prod_{i=2}^{d-1} \distNorm(x_{i} \given 0, \sigma)$. 
	Hence, if $X \distas \posteriorDensity$, then $\var(X_{i}) = \exp(1/2)$ for  $i \in \{2,\dots,d\}$.
	Since $X_{2}, \dots, X_{d}$ are identically distributed, we focus on the accuracy of the mean-field VI approximations to the distributions of $X_{1}$ and $X_{2}$. 

	\Cref{fig:High Dimensional Neal Funnel Diagnostics} shows that the lower bounds constructed using Barker are quite precise and 
	do not degrade as the dimension increases. (See \cref{section: sm-neal} for results for quantiles.)
	Moreover the reliability check passes for all $d$ (\cref{fig:High Dimensional Neal Funnel reliability checks} in \cref{section: sm-neal}). 
	The $\hat{k}$ diagnostic, however, is noisy and provides little insight.
	The Wasserstein-based mean and covariance upper bounds are orders-of-magnitude too large and unavailable for most $d > 10$ because $\hat{k}>0.7$.

	\subsection{Logistic Regression Model on Candy Power Ranking Dataset}
	\label{sec: experiment of candy classification}
	
	Next, we validate \diagnosticName on a real-world dataset\footnote{\url{https://www.kaggle.com/datasets/fivethirtyeight/the-ultimate-halloween-candy-power-ranking}}: each observation $(z_{n}, y_{n})$ represents a different kind of candy,
	where $z_{n} \in \reals^{11}$ are the features, and $y_{n} = 1$ if the candy is chocolate and 0 otherwise. 
	We use a logistic regression model with $y_{n} \distas \distBern(\operatorname{logit}^{-1}(\beta^{\top} z_{n}))$
	and $\beta_{i} \distas \distNorm(0, I)$. 

	\begin{figure}[t]
		\includegraphics[width=\linewidth,trim={0.5in 1.2in 1.3in 1.1in},clip]{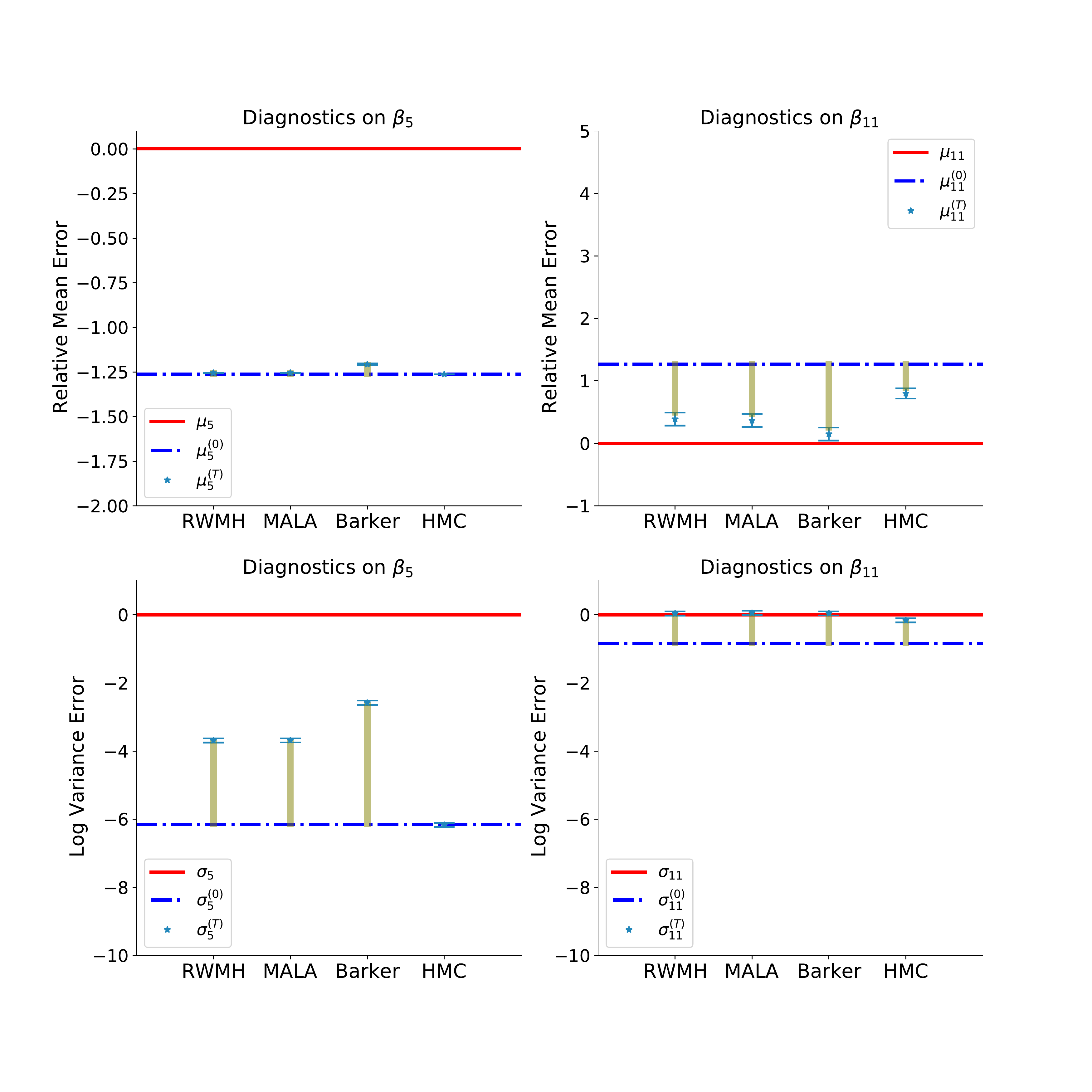}
		\caption{Diagnostics for logistic regression model on candy power ranking dataset for component $\beta_{5}$ and $\beta_{11}$. 
		Here $\mu_{i}$ and $\sigma_{i}$ denote, respectively, the mean and variance of $\beta_{i}$.
			To improve interpretability, we show the relative mean error ${(\mu_{i}^{( t ) }-\mu_{ i })}/{\sigma_{ i }}$
			and log variance error $2\log_{10}({\sigma_{i}^{(t)}}) - 2 \log_{10}({\sigma_{i}})$. 
			(Recall that $\mu_{i}$ and $\sigma_{i}$ are the $i$th component mean and standard deviation of the target $\posteriorDensity$.)
			The lengths of the gold lines give the corresponding relative mean and log variance error lower bounds.
		}
		\label{fig:TADAA for candy classification.}
	\end{figure}

	\begin{figure}[t]
		\centering
		\includegraphics[width=\linewidth]{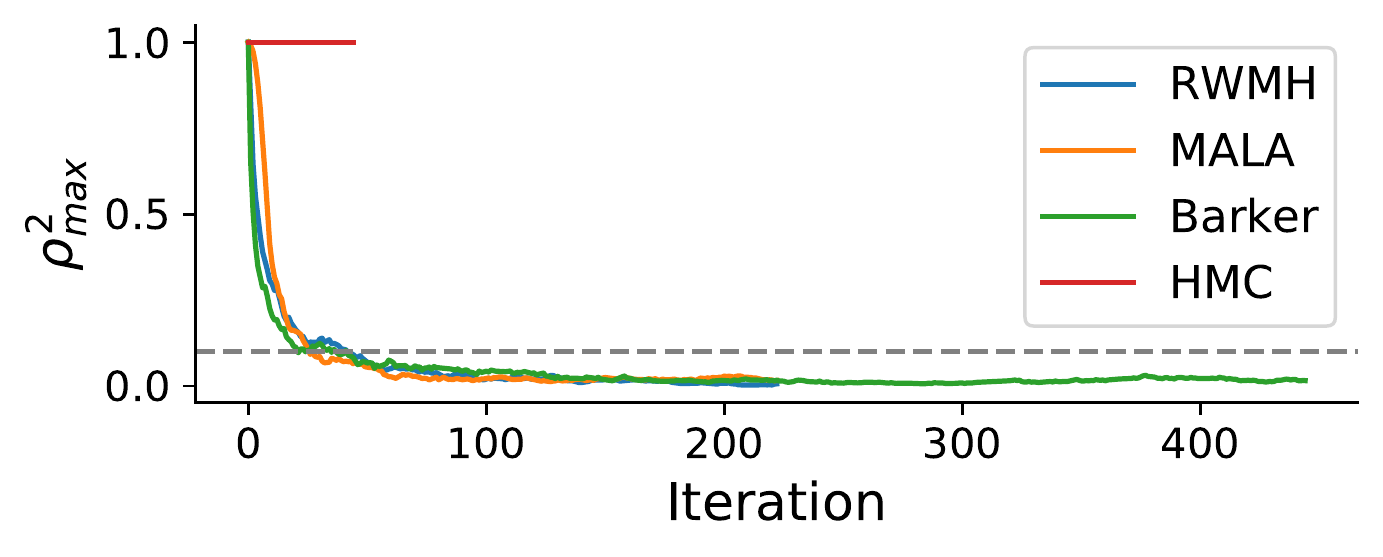}
		\caption{Reliability checks on \diagnosticName using different kernels applied to the mean-field VI approximation to logistic regression for candy classification.}
		\label{fig:Worst-case Reliability checks for candy classification diagnostics.}
	\end{figure}

	Focusing on Barker,
	\cref{fig:TADAA for candy classification.} shows that for $\beta_{5}$, the lower bound of the mean error is small but non-zero due
	to the fact that the variance is dramatically underestimated, leading to inefficient preconditioning (adaptive preconditioning did not significantly improve the lower bound).
	On the other hand, the variance error lower bound, while not tight, shows that the variance estimate is orders-of-magnitude too small. 
	For $\beta_{11}$, where the variance is not so poorly estimated, both mean and variance lower bounds are very accurate.
	\diagnosticName uses about 2\% as many gradient evaluations as the VI optimization ($4.3 \times 10^{4}$ versus $2 \times 10^{6}$), 
	validating the computational efficiency of our approach. 
	\Cref{fig:Worst-case Reliability checks for candy classification diagnostics.} demonstrates the usefulness of the reliability checks, which
	correctly show that HMC chains fail to mix (due to a smaller number of iterations), while the chains for other kernels mix well. 

	\subsection{Cancer Classification Using a Horseshoe Prior}
	\label{sec: Cancer Classification Using a Horseshoe Prior}
\begin{figure}[t]
		\includegraphics[width=\linewidth,trim={0.5in 1.2in 1.3in 1.1in},clip]{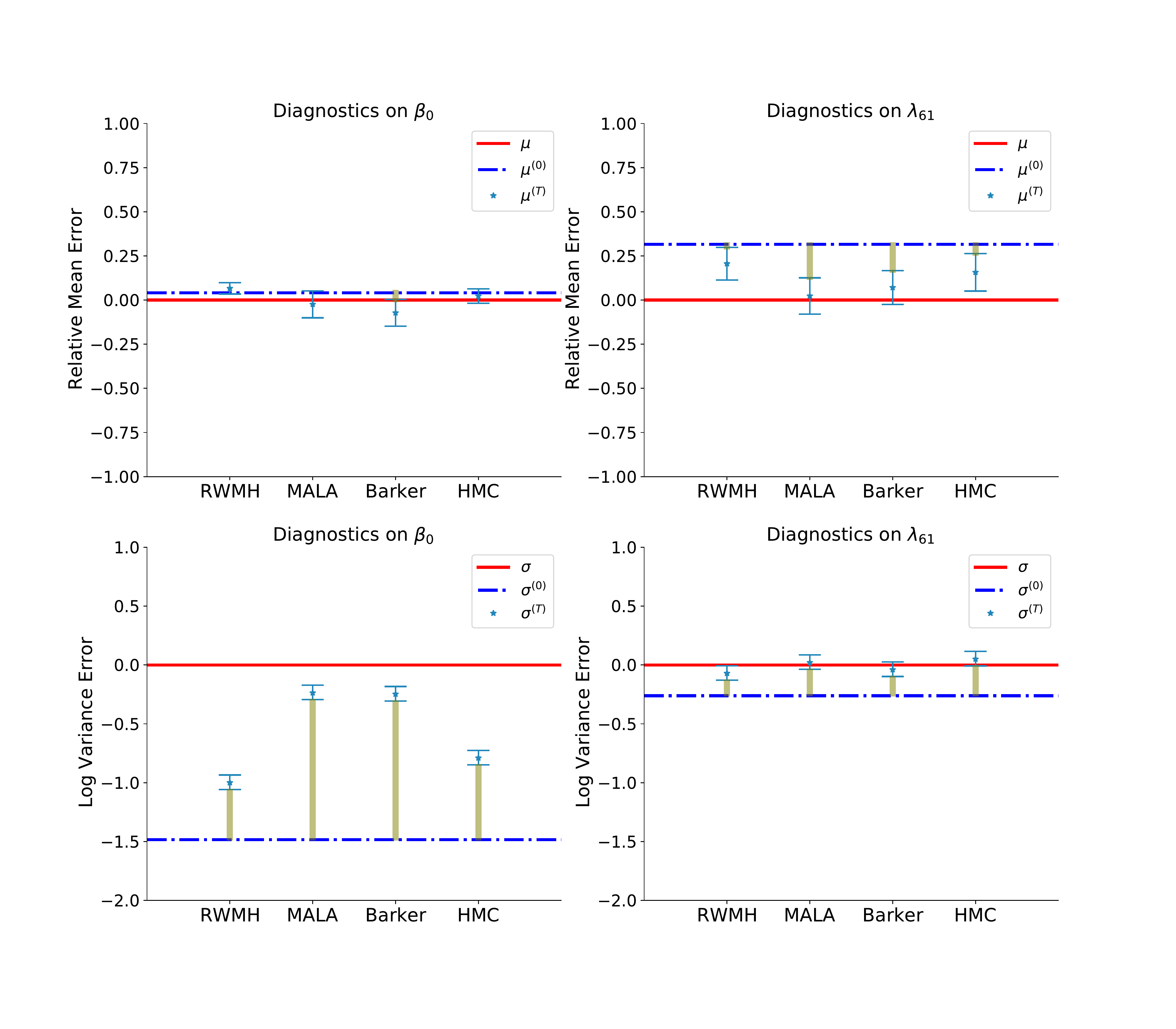}
		\caption{Diagnostics for $\beta_{0}$ and $\lambda_{61}$ for cancer classification using the horseshoe prior.
			See \cref{fig:TADAA for candy classification.} for further explanations.}
		\label{fig:TADAA for cancer classification}
	\end{figure}

\begin{figure}[t]
		\includegraphics[width=\linewidth]{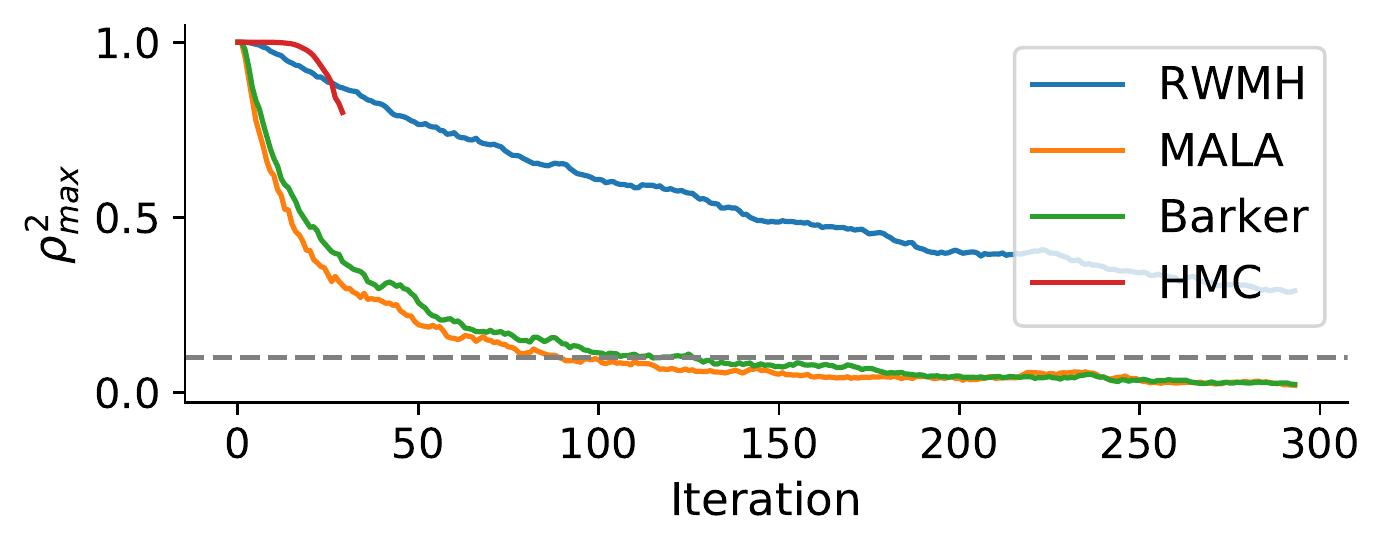}
		\caption{Reliability checks for \diagnosticName using different kernels applied to horseshoes prior applied to mean-field approximation for cancer classification.}
		\label{fig:Reliability checks for cancer classification}
	\end{figure}

	We now consider a more challenging, higher-dimensional application to predicting leukemia using 
	microarray data with $n = 71$ observations %
	and using 100 features chosen according to their $\chi^{2}$ scores \citep{ray2020chi}. 
	We use a logistic regression model with a sparsity-inducing horseshoe prior \citep{piironen2017sparsity}:
	\[
	\begin{aligned}
		y &\mid \beta \sim \distBern(\operatorname{logit}^{-1}(X \beta)),\\
		\beta_{j} &\mid \tau, \lambda, c \sim \distNorm(0, \tau^{2} \tilde{\lambda}_{j}^{2}), \\
		\lambda_{j} &\sim \mathrm{C}^{+}(0,1), \qquad \tau \sim \mathrm{C}^{+}\left(0, \tau_{0}\right), \\
		c^{2} &\sim \distInvGam(2,8),
	\end{aligned}
	\]
	where $y$ denotes the binary outcomes, $X \in \reals^{71 \times 100}$ is the features matrix, 
	$\tau >0 $ and $\lambda >0$ are global and local shrinkage parameters, 
	and $\tilde{\lambda}_{j}^{2} = c^{2} \lambda_{j}^{2} /(c^{2}+\tau^{2} \lambda_{j}^{2})$. 
	Hence, for our data the parameter dimensionality is $d = 203$. 
	We would expect the VI approximation to be poor due to multimodality of the model.
	Focusing on two representative parameters, \cref{fig:TADAA for cancer classification} shows that \diagnosticName using Barker correctly captures 
	that the mean estimate for $\lambda_{61}$ and variance estimate for $\beta_{0}$ are inaccurate. 
	On the other hand, for the accurate mean estimate for $\beta_{0}$ and variance estimate for $\lambda_{61}$, the lower bounds are zero or close to zero. 
	\diagnosticName uses about 28\% as many gradient evaluations as the VI optimization ($1.1 \times 10^{5}$ versus $4 \times 10^{5}$).
	In addition, the reliability checks shown in \cref{fig:Reliability checks for cancer classification} correctly reflect the worse
	performance of RWMH (due to $d$ being larger) and HMC (due to a smaller number of iterations).

	\subsection{Bayesian Neural Network on Candy Power Ranking Dataset}
	\label{sec: Bayesian Neural Network}
\begin{figure}
		\includegraphics[width=\linewidth]{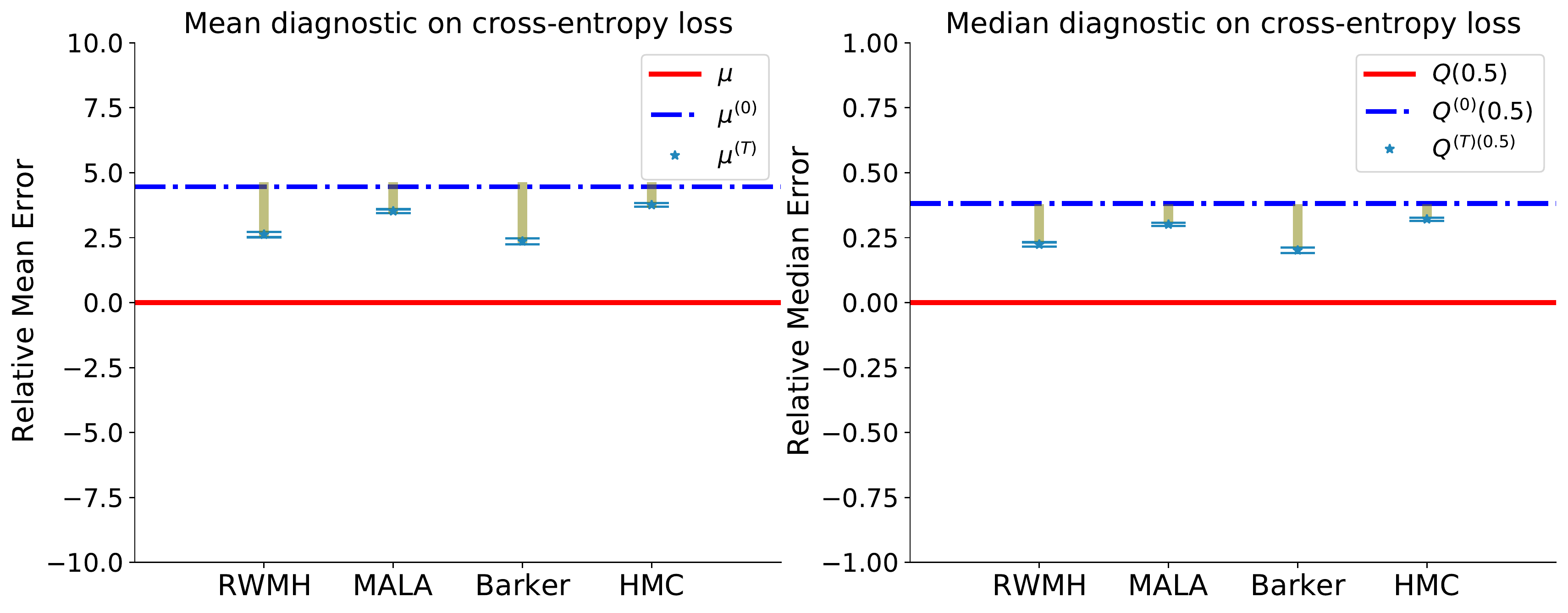}
		\caption{Diagnostics of  for Bayesian neural network model on candy power ranking dataset for component. $\mu$ and $Q(0.5)$ denote the mean and median for the cross-entropy loss.}
		\label{fig:TADAA for candy classification_bnn.}
	\end{figure}

	Finally, we validate \diagnosticName on a Bayesian neural network model, which is difficult for mean-field VI to approximate accurately \citep{izmailov2021bayesian}. 
	We use the same dataset described in \cref{sec: experiment of candy classification}, but now it is fitted with a two-layer Bayesian neural network with 5 units in each hidden layer: 
	\[
	y_{n} \distas \distBern\left(\operatorname{logit}^{-1}\left( \operatorname{tanh} \left( \operatorname{tanh} (\alpha^{\top} z_{n})\cdot \beta \right) \cdot \gamma \right)\right),
	\]
	where $\alpha \in \reals^{5 \times 11}$, $\beta \in \reals^{5 \times 5}$, $\gamma \in \reals^{5}$, $\alpha_{i,j} \distas \distNorm(0, 2)$, $\beta_{i,j} \distas \distNorm(0, 2)$ and $\gamma_{i} \distas \distNorm(0, 2)$. 
	Hence, $x = (\alpha, \beta, \gamma)$ and $d = 85$. 
	To quantify the classification quality for the Bayesian neural network, we can use the log loss (a.k.a., the cross-entropy loss)
	$
	\mathcal{L}(x) = -\frac{1}{N}\sum_{n=1}^{N}\log p(y_{n} \given  x).
	$
	Similar to the definitions for the marginal mean and median, we can also use the mean and median of the log loss 
	as the evaluation functional $\mcF$. 
	\Cref{fig:TADAA for candy classification_bnn.} shows that \diagnosticName correctly captures that the inaccuracy of the VI approximation:
	error lower bounds show that both the mean and median of the log loss are overestimated by mean-field VI.
	\diagnosticName uses about 12$\%$ as many gradient evaluations as the VI optimization ($8.4\times10^{4}$ versus $6.8\times10^{5}$).
	The reliability check in \cref{fig:Reliability checks for bnn} confirms the superiority of Barker in high-dimensional problems. 
	
	\section{CONCLUSION AND LIMITATIONS}
	
	Overall, our theory and experimental results show that \diagnosticName provides an efficient evaluation tool for variational inference,
	while also being applicable to other inexact posterior approximation methods like (integrated nested) Laplace approximations. 
	\diagnosticName provides precise information about specific functionals of interest such as means and standard deviations, rather than a check on just the overall quality of an 
	approximation -- which in practice is often quite poor. 
	However, there are a number of important limitations that users must keep in mind when employing \diagnosticName. 
	While we have developed the correlation check to guard against a poor diagnostic, the check can fail due to, e.g., multimodality.
	On the other hand, such failures are not unique to our diagnostic and can affect, e.g., diagnostics for MCMC too.
	One approach to guard against this possibility is to run the VI optimization multiple times with diverse initializations, as is usually recommended for MCMC. 
	Another limitation is that \diagnosticName is not appropriate in settings where it is only feasible to make a small number of passes over the entire dataset,
	so a method like stochastic variational inference \citep{hoffman2013stochastic} is used.
	Finally, it is possible that \cref{cond:monotonicity} could be violated. 
	Developing checks against this failure mode is an important direction for future work. 

\subsubsection*{Acknowledgments}
M.~Kasprzak was support by the European Union's Horizon 2020 research and innovation programme under the Marie Skłodowska-Curie grant agreement No.\ 101024264.
J.~H.~Huggins was partially supported by the National Institute of General Medical Sciences of the National Institutes of Health under grant number R01GM144963 as part of the Joint NSF/NIGMS Mathematical Biology Program. 
The content is solely the responsibility of the authors and does not necessarily represent the official views of the National Institutes of Health.

\twocolumn
\bibliography{reference_new}

\begin{thebibliography}{}

\bibitem[Agrawal et~al., 2020]{agrawal2020advances}
Agrawal, A., Sheldon, D.~R., and Domke, J. (2020).
\newblock {Advances in black-box VI: Normalizing flows, importance weighting,
  and optimization}.
\newblock {\em Advances in Neural Information Processing Systems},
  33:17358--17369.

\bibitem[Andrieu and Thoms, 2008]{andrieu2008tutorial}
Andrieu, C. and Thoms, J. (2008).
\newblock {A tutorial on adaptive MCMC}.
\newblock {\em Statistics and Computing}, 18(4):343--373.

\bibitem[Atchad{\'e} and Rosenthal, 2005]{atchade2005adaptive}
Atchad{\'e}, Y.~F. and Rosenthal, J.~S. (2005).
\newblock {On adaptive Markov chain Monte Carlo algorithms}.
\newblock {\em Bernoulli}, 11(5):815 -- 828.

\bibitem[Beskos et~al., 2013]{beskos2013optimal}
Beskos, A., Pillai, N., Roberts, G., Sanz-Serna, J.-M., and Stuart, A. (2013).
\newblock {Optimal tuning of the hybrid Monte Carlo algorithm}.
\newblock {\em Bernoulli}, 19(5A):1501--1534.

\bibitem[Bhatia et~al., 2022]{bhatia2022statistical}
Bhatia, K., Kuang, N.~L., Ma, Y.-A., and Wang, Y. (2022).
\newblock {Statistical and Computational Trade-offs in Variational Inference: A
  Case Study in Inferential Model Selection}.
\newblock {\em arXiv preprint arXiv:2207.11208}.

\bibitem[Billingsley, 2013]{billingsley2013convergence}
Billingsley, P. (2013).
\newblock {\em {Convergence of probability measures}}.
\newblock John Wiley \& Sons.

\bibitem[Bishop, 2006]{bishop2006pattern}
Bishop, C.~M. (2006).
\newblock {\em {Pattern recognition and machine learning}}, volume~4.
\newblock Springer.

\bibitem[Blei et~al., 2017]{blei2017variational}
Blei, D.~M., Kucukelbir, A., and McAuliffe, J.~D. (2017).
\newblock {Variational inference: A review for statisticians}.
\newblock {\em Journal of the American Statistical Association},
  112(518):859--877.

\bibitem[Brooks et~al., 2011]{brooks2011handbook}
Brooks, S., Gelman, A., Jones, G., and Meng, X.-L. (2011).
\newblock {\em {Handbook of Markov chain Monte Carlo}}.
\newblock CRC press.

\bibitem[Burda et~al., 2015]{burda2015importance}
Burda, Y., Grosse, R., and Salakhutdinov, R. (2015).
\newblock {Importance weighted autoencoders}.
\newblock {\em arXiv preprint arXiv:1509.00519}.

\bibitem[Chee and Toulis, 2018]{chee2017convergence}
Chee, J. and Toulis, P. (2018).
\newblock {Convergence diagnostics for stochastic gradient descent with
  constant learning rate}.
\newblock In {\em International Conference on Artificial Intelligence and
  Statistics}, pages 1476--1485. PMLR.

\bibitem[Craiu et~al., 2009]{craiu2009learn}
Craiu, R.~V., Rosenthal, J., and Yang, C. (2009).
\newblock {Learn from thy neighbor: Parallel-chain and regional adaptive MCMC}.
\newblock {\em Journal of the American Statistical Association},
  104(488):1454--1466.

\bibitem[Domke and Sheldon, 2018]{domke2018importance}
Domke, J. and Sheldon, D.~R. (2018).
\newblock {Importance weighting and variational inference}.
\newblock {\em Advances in Neural Information Processing Systems}, 31.

\bibitem[Gelman et~al., 1995]{gelman1995bayesian}
Gelman, A., Carlin, J.~B., Stern, H.~S., and Rubin, D.~B. (1995).
\newblock {\em {Bayesian data analysis}}.
\newblock Chapman and Hall/CRC.

\bibitem[Gelman et~al., 1997]{gelman1997weak}
Gelman, A., Gilks, W.~R., and Roberts, G.~O. (1997).
\newblock {Weak convergence and optimal scaling of random walk Metropolis
  algorithms}.
\newblock {\em The Annals of Applied Probability}, 7(1):110--120.

\bibitem[Gelman and Rubin, 1992]{gelman1992inference}
Gelman, A. and Rubin, D.~B. (1992).
\newblock {Inference from iterative simulation using multiple sequences}.
\newblock {\em Statistical Science}, pages 457--472.

\bibitem[Gomez-Rubio, 2020]{GomezRubio:2020:INLA}
Gomez-Rubio, V. (2020).
\newblock {\em {Bayesian inference with INLA}}.
\newblock CRC Press.

\bibitem[Gorham and Mackey, 2015]{mackey2016measuring}
Gorham, J. and Mackey, L. (2015).
\newblock {Measuring sample quality with Stein's method}.
\newblock {\em Advances in Neural Information Processing Systems}, 28.

\bibitem[Gorham and Mackey, 2017]{gorham2017measuring}
Gorham, J. and Mackey, L. (2017).
\newblock {Measuring sample quality with kernels}.
\newblock In {\em International Conference on Machine Learning}, pages
  1292--1301. PMLR.

\bibitem[Green et~al., 2015]{green2015}
Green, P.~J., {\L}atuszy\'nski, K., Pereyra, M., and Robert, C.~P. (2015).
\newblock {Bayesian computation: a summary of the current state, and samples
  backwards and forwards}.
\newblock {\em Statistics and Computing}, 25(4):835--862.

\bibitem[Hahn and Meeker, 2011]{hahn2011statistical}
Hahn, G.~J. and Meeker, W.~Q. (2011).
\newblock {\em Statistical intervals: a guide for practitioners}, volume~92.
\newblock John Wiley \& Sons.

\bibitem[Hoffman et~al., 2013]{hoffman2013stochastic}
Hoffman, M.~D., Blei, D.~M., Wang, C., and Paisley, J. (2013).
\newblock {Stochastic Variational Inference}.
\newblock {\em Journal of Machine Learning Research}, 14:1303--1347.

\bibitem[Huggins et~al., 2020]{huggins2020validated}
Huggins, J., Kasprzak, M., Campbell, T., and Broderick, T. (2020).
\newblock {Validated variational inference via practical posterior error
  bounds}.
\newblock In {\em International Conference on Artificial Intelligence and
  Statistics}, pages 1792--1802. PMLR.

\bibitem[Izmailov et~al., 2021]{izmailov2021bayesian}
Izmailov, P., Vikram, S., Hoffman, M.~D., and Wilson, A. G.~G. (2021).
\newblock {What Are Bayesian Neural Network Posteriors Really Like?}
\newblock In {\em Proceedings of the 38th International Conference on Machine
  Learning}, volume 139 of {\em Proceedings of Machine Learning Research},
  pages 4629--4640. PMLR.

\bibitem[Jordan et~al., 1999]{jordan1999introduction}
Jordan, M.~I., Ghahramani, Z., Jaakkola, T.~S., and Saul, L.~K. (1999).
\newblock {An introduction to variational methods for graphical models}.
\newblock {\em Machine Learning}, 37(2):183--233.

\bibitem[Kingma and Welling, 2013]{kingma2013auto}
Kingma, D.~P. and Welling, M. (2013).
\newblock {Auto-encoding variational Bayes}.
\newblock {\em arXiv preprint arXiv:1312.6114}.

\bibitem[Kucukelbir et~al., 2017]{kucukelbir2017automatic}
Kucukelbir, A., Tran, D., Ranganath, R., Gelman, A., and Blei, D.~M. (2017).
\newblock {Automatic differentiation variational inference}.
\newblock {\em Journal of Machine Learning Research}.

\bibitem[Livingstone and Zanella, 2022]{livingstone2019barker}
Livingstone, S. and Zanella, G. (2022).
\newblock {The Barker proposal: Combining robustness and efficiency in
  gradient-based MCMC}.
\newblock {\em Journal of the Royal Statistical Society. Series B, Statistical
  Methodology}, 84(2):496.

\bibitem[Mendenhall et~al., 2012]{mendenhall2012introduction}
Mendenhall, W., Beaver, R.~J., and Beaver, B.~M. (2012).
\newblock {\em Introduction to probability and statistics}.
\newblock Cengage Learning.

\bibitem[Metropolis et~al., 1953]{metropolis1953equation}
Metropolis, N., Rosenbluth, A.~W., Rosenbluth, M.~N., Teller, A.~H., and
  Teller, E. (1953).
\newblock {Equation of state calculations by fast computing machines}.
\newblock {\em The Journal of Chemical Physics}, 21(6):1087--1092.

\bibitem[Minka, 2001]{Minka:2001}
Minka, T.~P. (2001).
\newblock {Expectation propagation for approximate Bayesian inference}.
\newblock In {\em Uncertainty in Artificial Intelligence}.

\bibitem[Neal, 2011]{neal2011mcmc}
Neal, R. (2011).
\newblock {MCMC using Hamiltonian dynamics}.
\newblock {\em Handbook of Markov Chain Monte Carlo}, pages 113--162.

\bibitem[Neal, 2003]{neal2003slice}
Neal, R.~M. (2003).
\newblock {Slice sampling}.
\newblock {\em The Annals of Statistics}, 31(3):705--767.

\bibitem[Piironen and Vehtari, 2017]{piironen2017sparsity}
Piironen, J. and Vehtari, A. (2017).
\newblock {Sparsity information and regularization in the horseshoe and other
  shrinkage priors}.
\newblock {\em Electronic Journal of Statistics}, 11(2):5018--5051.

\bibitem[Ray et~al., 2020]{ray2020chi}
Ray, S., Alshouiliy, K., Roy, A., AlGhamdi, A., and Agrawal, D.~P. (2020).
\newblock {Chi-Squared Based Feature Selection for Stroke Prediction using
  AzureML}.
\newblock In {\em 2020 Intermountain Engineering, Technology and Computing
  (IETC)}, pages 1--6. IEEE.

\bibitem[Robert et~al., 2007]{robert2007bayesian}
Robert, C.~P. et~al. (2007).
\newblock {\em {The Bayesian choice: from decision-theoretic foundations to
  computational implementation}}, volume~2.
\newblock Springer.

\bibitem[Roberts and Rosenthal, 1998]{roberts1998optimal}
Roberts, G.~O. and Rosenthal, J.~S. (1998).
\newblock {Optimal scaling of discrete approximations to Langevin diffusions}.
\newblock {\em Journal of the Royal Statistical Society: Series B (Statistical
  Methodology)}, 60(1):255--268.

\bibitem[Roberts and Rosenthal, 2001]{Roberts:2001}
Roberts, G.~O. and Rosenthal, J.~S. (2001).
\newblock {Optimal scaling for various Metropolis-Hastings algorithms}.
\newblock {\em Statistical Science}, 16(4):351 -- 367.

\bibitem[Roberts and Rosenthal, 2004]{roberts2004general}
Roberts, G.~O. and Rosenthal, J.~S. (2004).
\newblock {General state space Markov chains and MCMC algorithms.}
\newblock {\em Probability Surveys}, 1:20--71.

\bibitem[Roberts and Rosenthal, 2007]{roberts2007coupling}
Roberts, G.~O. and Rosenthal, J.~S. (2007).
\newblock {Coupling and Ergodicity of Adaptive Markov Chain Monte Carlo
  Algorithms}.
\newblock {\em Journal of Applied Probability}, 44(2):458--475.

\bibitem[Roberts and Tweedie, 1996]{roberts1996exponential}
Roberts, G.~O. and Tweedie, R.~L. (1996).
\newblock {Exponential convergence of Langevin distributions and their discrete
  approximations}.
\newblock {\em Bernoulli}, pages 341--363.

\bibitem[Rosenthal, 2000]{rosenthal2000parallel}
Rosenthal, J.~S. (2000).
\newblock {Parallel computing and Monte Carlo algorithms}.
\newblock {\em Far East Journal of Theoretical Statistics}, 4(2):207--236.

\bibitem[Roy and Zhang, 2022]{roy2022convergence}
Roy, V. and Zhang, L. (2022).
\newblock {Convergence of position-dependent MALA with application to
  conditional simulation in GLMMs}.
\newblock {\em Journal of Computational and Graphical Statistics},
  (just-accepted):1--31.

\bibitem[Rue et~al., 2009]{Rue:2009}
Rue, H., Martino, S., and Chopin, N. (2009).
\newblock {Approximate Bayesian inference for latent Gaussian models by using
  integrated nested Laplace approximations}.
\newblock {\em Journal of the Royal Statistical Society: Series B (Statistical
  Methodology)}, 71(2):319 -- 392.

\bibitem[Sirignano and Spiliopoulos, 2020]{sirignano2020mean}
Sirignano, J. and Spiliopoulos, K. (2020).
\newblock {Mean field analysis of neural networks: A law of large numbers}.
\newblock {\em SIAM Journal on Applied Mathematics}, 80(2):725--752.

\bibitem[Solonen et~al., 2012]{solonen2012efficient}
Solonen, A., Ollinaho, P., Laine, M., Haario, H., Tamminen, J., and
  J{\"a}rvinen, H. (2012).
\newblock {Efficient MCMC for climate model parameter estimation: Parallel
  adaptive chains and early rejection}.
\newblock {\em Bayesian Analysis}, 7(3):715--736.

\bibitem[Stramer and Roberts, 2007]{stramer2007bayesian}
Stramer, O. and Roberts, G.~O. (2007).
\newblock {On Bayesian analysis of nonlinear continuous-time autoregression
  models}.
\newblock {\em Journal of Time Series Analysis}, 28(5):744--762.

\bibitem[Sznitman, 1991]{sznitman1991topics}
Sznitman, A.-S. (1991).
\newblock {\em {Topics in propagation of chaos}}.
\newblock Springer.

\bibitem[Vats and Knudson, 2021]{vats2021revisiting}
Vats, D. and Knudson, C. (2021).
\newblock {Revisiting the gelman--rubin diagnostic}.
\newblock {\em Statistical Science}, 36(4):518--529.

\bibitem[Vehtari et~al., 2021]{vehtari2021rank}
Vehtari, A., Gelman, A., Simpson, D., Carpenter, B., and B{\"u}rkner, P.-C.
  (2021).
\newblock {Rank-normalization, folding, and localization: an improved $\hat{R}$
  for assessing convergence of MCMC (with discussion)}.
\newblock {\em Bayesian Analysis}, 16(2):667--718.

\bibitem[Wada and Fujisaki, 2015]{wada2015stopping}
Wada, T. and Fujisaki, Y. (2015).
\newblock {A stopping rule for stochastic approximation}.
\newblock {\em Automatica}, 60:1--6.

\bibitem[Wainwright et~al., 2008]{wainwright2008graphical}
Wainwright, M.~J., Jordan, M.~I., et~al. (2008).
\newblock {Graphical models, exponential families, and variational inference}.
\newblock {\em Foundations and Trends{\textregistered} in Machine Learning},
  1(1--2):1--305.

\bibitem[Welandawe et~al., 2022]{Welandawe:2022:BBVI}
Welandawe, M., Andersen, M.~R., Vehtari, A., and Huggins, J.~H. (2022).
\newblock {Robust, Automated, and Accurate Black-box Variational Inference}.
\newblock {\em arXiv}, arXiv:2203.15945 [stat.ML].

\bibitem[Xing et~al., 2020]{xing2020distortion}
Xing, H., Nicholls, G., and Lee, J.~K. (2020).
\newblock {Distortion estimates for approximate Bayesian inference}.
\newblock In {\em Conference on Uncertainty in Artificial Intelligence}, pages
  1208--1217. PMLR.

\bibitem[Yao et~al., 2018]{yao2018yes}
Yao, Y., Vehtari, A., Simpson, D., and Gelman, A. (2018).
\newblock {Yes, but did it work?: Evaluating variational inference}.
\newblock In {\em International Conference on Machine Learning}, pages
  5581--5590. PMLR.

\end{thebibliography}

\clearpage
\appendix
\onecolumn
\aistatstitle{Supplementary Materials for\\A Targeted Accuracy Diagnostic for Variational Approximations}
\section{METROPOLIS-HASTING KERNELS} 
\label{sec:MH-kernels}

We briefly summarize the kernels used in our experiments. 
Throughout this section let $x \in \reals^{d}$ denote the current state, 
$h \in \reals_{+}$ the step size, and $G \in \reals^{d \times d}$ a positive semi-definition preconditioning matrix. 

\paragraph{Random Walk Metropolis}
The (pre-conditioned) random walk Metropolis (RWMH) \citet{metropolis1953equation} with and step size $h$ has proposal kernel 
\[
Q^{RW}(x, \dee y) = \distNorm(\dee y \given x, h G).
\]

\paragraph{Metropolis-adjusted Langevin Algorithm}
In the (pre-conditioned) Metropolis-adjusted Langevin algorithm (MALA) \citep{roberts1996exponential,stramer2007bayesian}
has proposal kernel 
\[
Q^{L}(x, \dee y) = \distNorm\left(\dee y \biggm| x + \frac{h}{2} G \nabla \log \posteriorDensity(x), h G\right).
\]

\paragraph{Barker Proposal}
Let $C$ denote the Cholesky factor of $G$, set $\tau_{i}^{2} = hG_{ii}$, let $\mu(\cdot)_{\tau_{i}}$ be the probability density function of $\mathcal{N}(0, \tau_{i}^{2})$, and let $c_i(x)=(\nabla \log \pi(x) \cdot C^T)_i$. 
For a proposal state $y \in \reals^{d}$, let $z_i=((C^\top) ^{-1}(y-x))_i$. 
The (pre-conditioned) Barker proposal \citet{livingstone2019barker} for the $i$th coordinate is 
\[
Q_{i}^{B}(x, \dee y_{i})=2 \frac{\mu_{\tau_{i}}(y_{i}-x_{i})}{1+e^{-z_{i}c_{i}(x)}} \dee y_{i},
\]
and the full Barker kernel $Q^{B}$ is 
\[
Q^{B}(x, \dee y)=\prod_{i=1}^{d} Q_{i}^{B}\left(x, \dee y_{i}\right).
\]

\paragraph{Hamiltonian Monte Carlo}
Hamiltonian Monte Carlo (HMC) \citep{neal2011mcmc} is defined on an extended state space with the random momentum vector $\eta \distind \distNorm(0, G)$,
yielding the joint density
\[
\bar\pi(x, \eta) \propto \exp \left\{\log(\posteriorDensity(x))-\frac{1}{2} \eta^{\top}G\eta\right\}.
\]
HMC updates a new state using the leapfrog integrator, which proceeds according to the updates
\[
\eta(t+h / 2) &=\eta(t)+(h/ 2) \nabla_{x} \log\posteriorDensity\left(x(t)\right)\\
x(t+h) &=x(t)+h G \eta(t+h / 2)\\
\eta(t+h) &=\eta(t+h/ 2)+(h/ 2)  \nabla_{x} \log\posteriorDensity\left(x(t+h)\right).
\]
Taking $x(0)=x$, the new proposal state is $y=x(Lh)$ with $\eta(0) \sim \distNorm(0, G)$.

\section{PROOFS}
\label{proofs}
For readability, we write the proposal kernel as $K(x, h, \dee y) = K_{\log h}(x, \dee y)$, so
\[
X_{j}^{(t+1)} \sim K\left(X_{j}^{(t)}, \CurrentStepsize, \cdot\right).
\]
Or, equivalently, 
\[
X_{j}^{(t+1)}= \begin{cases}Y_{j}^{(t)}, & \quad \text { with probability } \alpha_{j}^{(t)} \\ X_{j}^{(t)}, & \quad \text { with probability } 1-\alpha_{j}^{(t)},
\end{cases}
\]
with $Y_{j}^{(t)}$ being the proposal state based on the proposal distribution $Q_{\CurrentStepsize}(X_{j}^{(t)}, \cdot)$ and
\[
\begin{aligned}
	\alpha_{j}^{(t)} &\defas g(X_{j}^{(t)}, Y_{j}^{(t)},\CurrentStepsize)\\
	& \defas\min \left\{1, \frac{\posteriorDensity\left(Y_{j}^{(t)}\right) q_{\CurrentStepsize}\left(X_{j}^{(t)},Y_{j}^{(t)}\right)}{\posteriorDensity\left(X_{j}^{(t)}\right) q_{\CurrentStepsize}\left(Y_{j}^{(t)}, X_{j}^{(t)}\right)}\right\},
\end{aligned}
\]
where $q_{h}(x,\cdot)$ is the probability density of $Q_{h}(x, \cdot)$ and $\CurrentStepsize$ is the step size at time $t$.
If we treat the proposed states $\CurrentProposals$ as additional random variables, we can write 
\[
\left(X_{j}^{(t+1)}, Y_{j}^{(t+1)} \right) \sim T(X_{j}^{(t)}, Y_{j}^{(t)}, \CurrentStepsize, \cdot, \cdot)
\]
for an appropriate Markov kernel $T(x,y,h,\dee x', \dee y')$.
Now define the empirical distribution of $N$ particles $X_{j}^{(t)}$ at $t$th iteration as 
\[
\EmpiricalMeasureX \defas \frac{1}{N} \sum_{j=1}^{N} \delta_{X_{j}^{(t)}}.
\]
Similarly, we can define the empirical distribution of $N$ particles $(X_{j}^{(t)}, Y_{j}^{(t)})$ at the $t$th iteration as
\[
\EmpiricalMeasureXY \defas \frac{1}{N} \sum_{i=1}^{N} \delta_{(X_{i}^{(t)}, Y_{i}^{(t)})}.
\]

Taking the initial step size $h_{N}^{(0)}$ as fixed, we can rewrite the $t$th step size $\CurrentStepsize$ update equation given in \cref{eq:adaptive step size}) as
\[
\CurrentStepsize &=\PreviousStepsize \exp \left\{ \frac{1}{t} \left(\frac{1}{N} \sum_{j=1}^{N} \alpha_{j}^{(t-1)} - \bar{\alpha}_{*}\right) \right\}.
\]
Letting $\bar{h}^{(0)} = h_{N}^{(0)}$, $\bar{\nu}^{(0)} = \initialDensity$ and 
$\bar{\xi}^{(0)}(\dee x, \dee y)=Q_{ h_{N}^{(0)}}(x,  \dee y)\initialDensity(\dee x)$, then we can recursively define $\bar{h}^{(t)}$, $\bar{\nu}^{(t)}$ and $\bar{\xi}^{(t)}$ by 
\[
\bar{h}^{(t)} &\defas \LimitingStepsizePre  \exp \left( \frac{1}{\sqrt{t}}\langle g(\cdot, \cdot, \bar{h}^{(t-1)}), \bar{\xi}^{(t-1)} \rangle-\bar{\alpha}_{*}\right),\\
\bar{\nu}^{(t)} &\defas \int K(x, \bar{h}^{(t-1)},\cdot) \bar{\nu}^{(t-1)}(\dee x),\\
\bar{\xi}^{(t)} &\defas \int T(x, y, \bar{h}^{(t-1)},\cdot, \cdot) \bar{\xi}^{(t-1)}(\dee x, \dee y),
\]\
where 
\[
\begin{aligned}
	\langle g(\cdot, \cdot, \bar{h}^{(t-1)}), \bar{\xi}^{(t-1)} \rangle 
	&\defas \EE_{\left( X,Y\right) \sim \bar{\xi}^{(t-1)}} \left[ g\left(X,Y, \bar{h}^{(t-1)}\right)\right]\\
	& \defas \int g\left(x,y, \bar{h}^{(t-1)}\right) \bar{\xi}^{(t-1)}\left( \dee x, \dee y\right).
\end{aligned}
\]

Let $\mathcal{P}( S)$ denote the set of probability measures on the measurable space $(S, \mathcal{B})$. 
Let $\Xi_{N}^{(t)}$ and $\posterior_{N}^{(t)}$ denote the probability distributions of the respective empirical measures $\EmpiricalMeasureXY$ and $\EmpiricalMeasureX$, so $\EmpiricalMeasureXY \in \mathcal{P}(\mathcal{P}(\mathbb{R}^{d}))$ and $\posterior_{N}^{(t)} \in \mathcal{P}(\mathcal{P}(\mathbb{R}^{d} \times \mathbb{R}^{d}))$ respectively.

We will make use of the following two results. 
\begin{theorem}\citep{sznitman1991topics}
	\label{propagation of chaos theorem}
	$\{P_{N}^{(t)}\}$ is $\LimitingMeasureXt$-chaotic if and only if
	\[
	P_{N}^{(t)} \circ (\EmpiricalMeasureX)^{-1} \xrightarrow{d} \delta(\LimitingMeasureXt)
	\]
	in $\mathcal{P}(\mathcal{P}(S))$.
\end{theorem}

\begin{theorem}
	\label{continuous and bounded mapping}
	If $X, X_{n}: \Omega \to \mathbb{R}^{d}$ are random variables such that $X_{n} \xrightarrow{p} X$, then $f(X_{n}) \stackrel{\mathrm{L}^{1}}{\longrightarrow} f(X)$ for all $f \in \mathrm{C}_{b}(\mathbb{R}^{d})$.
\end{theorem}

\subsection{Proof of \cref{propagation of chaos for x}}

\Cref{propagation of chaos for x} will follow from a series of propositions and lemmas, which we prove in the subsequent subsections. 

\begin{proposition}
	\label{chaotic for the coupling}
	Under Assumption \ref{assumption}, if $(X_{1:N}^{(t-1)}, Y_{1:N}^{(t-1)})$ is $\bar{\xi}^{(t-1)}$-chaotic and $\PreviousStepsize \convp \LimitingStepsizePre$,
	then $(X_{1:N}^{(t)}, Y_{1:N}^{(t)})$ is $\LimitingMeasureXY$-chaotic.
\end{proposition}

\begin{proposition}
	\label{prop:convergence of step size}
	Under Assumptions \ref{assumption1} and \ref{assumption2}, if $(X_{1:N}^{(t-1)}, Y_{1:N}^{(t-1)})$ is $\LimitingMeasureXYPre$-chaotic and $\PreviousStepsize \convp \LimitingStepsizePre$, then $\CurrentStepsize \convp \LimitingStepsize$.
\end{proposition}
Note that $\big\{(X_{j}^{(0)}, Y_{j}^{(0)}) \big\}_{i=1}^{N}$ is $\bar{\xi}^{(0)}$-chaotic since the samples are \iid and $h_{N}^{(0)}=\bar{h}^{(0)}$ is fixed. By induction, it follows from \cref{chaotic for the coupling} and \cref{prop:convergence of step size} that for any $t \in \nats$, $\big\{(\CurrentSingleSample, \CurrentSingleProposal) \big\}_{i=1}^{N}$ is $\LimitingMeasureXY$-chaotic and $h_{k}^{N} \convp \LimitingStepsize$.

The following lemma implies that there is subsequence $\{\Pi_{N_{l}}^{(t)}\}_{l}$ that weakly converges.
\begin{lemma}
	\label{lemma:tightness lemma for X}
	Under Assumption \ref{assumption3}, for any fixed $t \in \nats$, $\{\posterior_{N}^{(t)}\}_{N}$ is tight and relatively compact.
\end{lemma}

\begin{lemma}
	\label{lemma:subsequence distribution convergence lemma for X}
	Under Assumption \ref{assumption}, if $X_{1:N}^{(t-1)}$ is $\bar{\nu}_{t-1}$-chaotic, $\PreviousStepsize \convp \LimitingStepsizePre$, 
	and $\{\posterior_{N_{l}}^{(t)}\}_{l}$ is a convergent subsequence of $\{\posterior_{N}^{(t)}\}_{N}$ with limit $\posterior^{(t)}$, then $\posterior^{(t)}$ is a Dirac measure concentrated on $\LimitingMeasureXt$.
\end{lemma}

\begin{lemma}
	\label{lemma:chaotic for x}
	Under assumptions and conditions of \cref{lemma:subsequence distribution convergence lemma for X}, $\CurrentSamples$ is $\LimitingMeasureXt$-chaotic.
\end{lemma}
We also know that $X_{1:N}^{(0)}$ is $\bar{\nu}^{(0)}$-chaotic (due to the fact that the $X_{1:N}^{(0)}$ are \iid), and $h_{k}^{N} \convp \LimitingStepsize$ for any $t \in \nats$.
Hence, by recursively using \cref{lemma:chaotic for x}, we conclude that $\CurrentSamples$ is $\LimitingMeasureXt$-chaotic for any $t \in \nats$.

\subsection{Proof of \cref{chaotic for the coupling}}
To prove \cref{chaotic for the coupling}, we will first show that the probability distribution $\{\Xi_{N}^{(t)}\}_{N}$ of empirical measure $\{\EmpiricalMeasureXY\}_{N}$ is tight and relatively compact, which guarantees that there exists convergent subsequence $\{\Xi_{N_{l}}^{(t)}\}_{l}$. Then we will prove that every convergent subsequence $\{\Xi_{N_{l}}^{(t)}\}_{l}$ has the same limit $\Xi^{(t)}$ that is a Dirac measure concentrated on $\LimitingMeasureXY$.

\begin{lemma}
	\label{tightness lemma}
	Under Assumption \ref{assumption3}, for any fixed $t \in \nats$, $\{\Xi_{N}^{(t)}\}_{N}$ is tight and relatively compact.
\end{lemma}

\begin{lemma}
	\label{lemma:subsequence distribution convergence lemma}
	Under Assumption \ref{assumption}, suppose $(X_{1:N}^{(t-1)}, Y_{1:N}^{(t-1)})$ is $\LimitingMeasureXYPre$-chaotic and $\PreviousStepsize \convp \LimitingStepsizePre$. Let  $\{\Xi_{N_{l}}^{(t)}\}_{l}$ be a convergent subsequence of $\{\Xi_{N}^{(t)}\}_{N}$ with limit $\Xi^{(t)}$. Then $\Xi^{(t)}$ is a Dirac measure concentrated on $\LimitingMeasureXY$.
\end{lemma}
The proofs of \cref{tightness lemma} and \ref{lemma:subsequence distribution convergence lemma} are deferred to \cref{Proof of Lemma tightness lemma} and \ref{Proof of Lemma subsequence distribution convergence lemma}. Relative compactness of $\{\Xi_{N}^{(t)}\}_{N}$ implies that there is subsequence $\{\Xi_{N_{l}}^{(t)}\}_{l}$ which weakly converges. In this proof, with a slight abuse of notation, we use $\{\Xi_{N}^{(t)}\}_{N}$ to represent the convergent subsequence $\{\Xi_{N_{l}}^{(t)}\}_{l}$.
By \cref{tightness lemma} and \cref{lemma:subsequence distribution convergence lemma}, we have
\[
\EmpiricalMeasureXY \convd \LimitingMeasureXY.
\]
\cref{propagation of chaos theorem} guarantees that $(X_{1:N}^{(t)}, Y_{1:N}^{(t)})$ is $\LimitingMeasureXY$-chaotic.

\subsubsection{Proof of \cref{tightness lemma}}
\label{Proof of Lemma tightness lemma}

The proof is similar to that of Lemma 2.2 in \citet{sirignano2020mean}.
For a fixed $t \in \nats$, $\forall \epsilon>0$, we want to show that there is a compact subset $K$ of $\mathcal{P}(\mathbb{R}^{d} \times \mathbb{R}^{d})$ such that 
\[
\sup _{N \in \mathbb{N}} \mathbb{P}\left[\EmpiricalMeasureXY \notin K\right]< \epsilon.
\]
For each $L>0$, define $K_{L}=[-L, L]^{2d}$, and by Assumption \ref{assumption3}, there exists $C$ such that
\[
\begin{aligned}
	\EE\left[\EmpiricalMeasureXY\left(\reals^{2d}  \backslash K_{L}\right)\right] & \leq \frac{1}{N} \sum_{j=1}^{N} \mathbb{P}\left[\left\| \left(\CurrentSingleSample, \CurrentSingleProposal\right)\right\| \geq L\right]  \\
	&= \mathbb{P}\left[\left\| \left(\CurrentSingleSample, \CurrentSingleProposal\right)\right\| \geq L\right] \\
	& \leq \frac{\EE \left[ \left\| \left(\CurrentSingleSample, \CurrentSingleProposal\right)\right\| \right]}{L}\\
	& \leq \frac{\EE \left[ \left\| \CurrentSingleSample\right\| + \left\| \CurrentSingleProposal\right\|\right]}{L} \\
	&= \frac{C}{L}.
\end{aligned}
\]
For $L > 0$, define the compact set
\[
\hat{K}_{L}=\overline{\left\{\nu: \nu\left(\mathbb{R}^{2d} \backslash K_{(L+j)^{2}}\right)<\frac{1}{\sqrt{L+j}} \text { for all } j \in \mathbb{N}\right\}}.
\]
Observe that
\[
\begin{aligned}
	\mathbb{P}\left[\EmpiricalMeasureXY \notin \hat{K}_{L}\right] & \leq \sum_{j=1}^{\infty} \mathbb{P}\left[\EmpiricalMeasureXY\left(\mathbb{R}^{2d} \backslash K_{(L+j)^{2}}\right)>\frac{1}{\sqrt{L+j}}\right] \leq \sum_{j=1}^{\infty} \frac{\EE\left[\EmpiricalMeasureXY\left(\mathbb{R}^{2d} \backslash K_{\left((L+j)^{2}\right)}\right)\right] }{1 / \sqrt{L+j}} \\
	& \leq \sum_{j=1}^{\infty} \frac{C}{(L+j)^{2} / \sqrt{L+j}} \leq \sum_{j=1}^{\infty} \frac{C}{(L+j)^{3 / 2}},
\end{aligned}
\]
so $\lim_{L \to \infty} \mathbb{P}[\EmpiricalMeasureXY\notin \hat{K}_{L}] = 0$. 
Thus, the sequence $(\EmpiricalMeasureXY)_{N \in \nats}$ is tight, which implies it is relatively compact.

\subsubsection{Proof of \cref{lemma:subsequence distribution convergence lemma}}
\label{Proof of Lemma subsequence distribution convergence lemma}

For each $f \in C_{b}(\mathbb{R}^{d} \times \mathbb{R}^{d} )$, define the map $F : \mathcal{P}(\mathbb{R}^{d} \times \mathbb{R}^{d})\rightarrow \mathbb{R}_{+}$ by
\[
F(\xi)= \left|\left\langle f, \xi \right\rangle-\left\langle f,\LimitingMeasureXY\right\rangle \right|.
\]
We would like to show that $\lim_{N \to \infty} \EE_{\xi \sim \Xi_{N}^{(t)}}[F(\xi)] = 0$. Together with the fact that subsequence $\{\Xi_{N}^{(t)}\}_{N}$ converges in distribution to $\Xi^{(t)}$, we would have $\EE_{\xi \sim \Xi^{(t)}}[F(\xi)] = 0$. Then we can conclude that $\Xi^{(t)}$ is a Dirac measure concentrated on $\LimitingMeasureXY$.

Letting $\langle f, \EmpiricalMeasureXYTilde \rangle = \frac{1}{N}\sum_{j=1}^{N}\langle f, T(\CurrentSingleSamplePre, \CurrentSingleProposalPre, \LimitingStepsizePre, \cdot, \cdot)\rangle$, 
we have 
\[
\label{upper bounds}
\begin{aligned}
	\EE_{\xi \sim \Xi_{N}^{(t)}}\left[F(\xi)\right] &=\EE\left[F\left(\EmpiricalMeasureXY\right)\right]\\
	& = \EE\left[\bigg| \left\langle f,\EmpiricalMeasureXY\right\rangle-\left\langle f,\LimitingMeasureXY\right\rangle\bigg|\right]\\
	&= \EE\left[\bigg| \left\langle f, \EmpiricalMeasureXY\right\rangle-\left\langle f, \EmpiricalMeasureXYTilde\right\rangle+\left\langle f, \EmpiricalMeasureXYTilde\right\rangle-\left\langle f,\LimitingMeasureXY\right\rangle \bigg| \right]\\
	& \leq  \EE\left[\bigg| \left\langle f, \EmpiricalMeasureXY\right\rangle-\left\langle f, \EmpiricalMeasureXYTilde\right\rangle \bigg|\right] +  \EE\left[\bigg|\left\langle f, \EmpiricalMeasureXYTilde\right\rangle-\left\langle f,\LimitingMeasureXY\right\rangle \bigg| \right],
\end{aligned}
\]
We will prove that $\lim_{N \to \infty}\EE_{\xi \sim \Xi_{N}^{(t)}}\left[F(\xi)\right] = 0$ by the following two lemmas:
\begin{lemma}
	\label{lemma: first upper bounds}
	Under the assumptions of \cref{lemma:subsequence distribution convergence lemma}, we have
	\[
	\lim_{N \to \infty} \EE\left[\bigg| \left\langle f, \EmpiricalMeasureXY\right\rangle-\left\langle f, \EmpiricalMeasureXYTilde\right\rangle \bigg|\right] = 0.
	\]
\end{lemma}

\begin{lemma}
	\label{lemma: second upper bounds}
	Under the assumptions of \cref{lemma:subsequence distribution convergence lemma}, we have
	\[
	\label{bounding term 2}
	\lim_{N \to \infty} \EE\left[\bigg|\left\langle f, \EmpiricalMeasureXYTilde\right\rangle-\left\langle f,\LimitingMeasureXY\right\rangle \bigg| \right]
	= 0.
	\]
\end{lemma}

By \cref{lemma: first upper bounds} and \ref{lemma: second upper bounds}, we have
\[
\lim_{N \to \infty} \EE_{\Xi_{N}^{(t)}}[F(\xi)] = 0.
\]
Since $F$ is uniformly bounded, we have
\[
\lim_{N \to \infty} \EE_{\Xi_{N}^{(t)}}[F(\xi)]  = \EE_{\Xi^{(t)}}[F(\xi)] = 0.
\]
Since this holds true for any $f \in C_{b}(\mathbb{R}^{d} \times \mathbb{R}^{d})$, using the fact that $F$ is nonnegative and by Theorem 5.1 from \citet{billingsley2013convergence}, we have that $\Xi_{N}^{(t)}$ has a limit point that is a Dirac measure concentrated on $\LimitingMeasureXY$.

\subsubsection{Proof of \cref{lemma: first upper bounds}}

	Let $\Filtrationcurrent$ be the $\sigma$-algebra generated by $X_{1:N}^{(1:m)}$ and $Y_{1:N}^{(1:m)}$, for $i=1,2,...,N$ and $m=1,2,...,t$.\\
	Note that
	\[
	\label{term 1}
	\begin{aligned}
		&\EE\left[\bigg| \left\langle f, \EmpiricalMeasureXY\right\rangle-\left\langle f, \EmpiricalMeasureXYTilde\right\rangle \bigg| \mid\FiltrationcurrentPre\right] \\
		& = \EE\left[\bigg|\frac{1}{N} \sum_{j=1}^{N} f\left(\CurrentSingleSample,  \CurrentSingleProposal\right)-\frac{1}{N}\langle f, T\left(\CurrentSingleSamplePre, \CurrentSingleProposalPre,\LimitingStepsizePre, \cdot, \cdot \right) \rangle \bigg|\mid\FiltrationcurrentPre\right]\\
		& \leq \EE\left [\bigg|\frac{1}{N} \sum_{j=1}^{N} f\left(\CurrentSingleSample,  \CurrentSingleProposal \right)-\frac{1}{N} \sum_{i=1}^{N} \langle f, T\left(\CurrentSingleSamplePre,  \CurrentSingleProposalPre, \PreviousStepsize, \cdot, \cdot \right ) \rangle \bigg|  \mid\FiltrationcurrentPre\right ] \\
		&+ \EE\left [\bigg| \frac{1}{N} \sum_{j=1}^{N} \langle f, T \left(\CurrentSingleSamplePre,  \CurrentSingleProposalPre,\PreviousStepsize, \cdot, \cdot \right) \rangle-\frac{1}{N} \sum_{i=1}^{N} \langle f, T \left(\CurrentSingleSamplePre,  \CurrentSingleProposalPre, \LimitingStepsizePre, \cdot, \cdot \right) \rangle \bigg| \mid\FiltrationcurrentPre\right ] \\
		&= \EE \left [\bigg| \frac{1}{N} \sum_{j=1}^{N} f\left(\CurrentSingleSample,  \CurrentSingleProposal\right)-\frac{1}{N} \sum_{i=1}^{N} \langle f, T \left(\CurrentSingleSamplePre,  \CurrentSingleProposalPre, \PreviousStepsize, \cdot, \cdot \right) \rangle \bigg|\mid\FiltrationcurrentPre\right ]  \\
		&+ \bigg| \frac{1}{N} \sum_{j=1}^{N} \langle f, T \left(\CurrentSingleSamplePre,  \CurrentSingleProposalPre, \PreviousStepsize, \cdot, \cdot \right) \rangle-\frac{1}{N} \sum_{j=1}^{N} \langle f, T \left(\CurrentSingleSamplePre,  \CurrentSingleProposalPre, \LimitingStepsizePre, \cdot, \cdot \right) \rangle \bigg|.
	\end{aligned}
	\]
	
	Since $f \in C_{b}(\mathbb{R}^{d} \times \mathbb{R}^{d})$, there exists $M$ such that for any $i=1,2,\ldots N$,  we have
	\begin{equation}
		\var\left[f(\CurrentSingleSample,  \CurrentSingleProposal) \mid\FiltrationcurrentPre \right] \leq M < \infty.
	\end{equation}
	It follows that
	\[
	\begin{aligned}
		&\EE\left[\bigg| \frac{1}{N} \sum_{j=1}^{N} f\left(\CurrentSingleSample,  \CurrentSingleProposal\right)-\frac{1}{N} \sum_{j=1}^{N} \langle f, T\left(\CurrentSingleSamplePre, \CurrentSingleProposalPre,\PreviousStepsize, \cdot, \cdot \right) \rangle\bigg| \mid\FiltrationcurrentPre\right] \\
		&= \EE\left[\bigg| \frac{1}{N}\sum_{j=1}^{N} f\left(\CurrentSingleSample ,  \CurrentSingleProposal\right) - \frac{1}{N}\EE[ \sum_{j=1}^{N} f\left(\CurrentSingleSample,  \CurrentSingleProposal\right)|\FiltrationcurrentPre] \bigg| \mid \FiltrationcurrentPre\right]\\
		& \leq \EE^{1/2} \left[\left( \frac{1}{N}\sum_{j=1}^{N} f\left(\CurrentSingleSample,  \CurrentSingleProposal\right) - \frac{1}{N}\EE[ \sum_{j=1}^{N} f\left(\CurrentSingleSample,  \CurrentSingleProposal\right)\mid\FiltrationcurrentPre] \right)^{2}|\FiltrationcurrentPre \right]\\
		& = \var^{1/2}\left[\frac{1}{N}\sum_{j=1}^{N} f\left(\CurrentSingleSample ,  \CurrentSingleProposal\right) \mid \FiltrationcurrentPre\right]\\
		& \leq \left(\frac{M}{N}\right)^{1/2}.
	\end{aligned}
	\]
	The last inequality holds true because $(\CurrentSingleSample, \CurrentSingleProposal)$ are independent given $\FiltrationcurrentPre$. Therefore, we have 
	\[
	\lim_{N \to \infty} \EE\left[\bigg| \frac{1}{N} \sum_{j=1}^{N} f\left(\CurrentSingleSample,  \CurrentSingleProposal\right)-\frac{1}{N} \sum_{j=1}^{N} \langle f, T\left(\CurrentSingleSamplePre, \CurrentSingleProposalPre,\PreviousStepsize, \cdot, \cdot \right) \rangle\bigg| \mid\FiltrationcurrentPre\right] = 0.
	\]
	And $\EE\left[\bigg| \frac{1}{N} \sum_{i=1}^{N} f\left(\CurrentSingleSample,  \CurrentSingleProposal\right)-\frac{1}{N} \sum_{i=1}^{N} \langle f, T\left(\CurrentSingleSamplePre, \CurrentSingleProposalPre,\PreviousStepsize, \cdot, \cdot \right) \rangle\bigg| \mid\FiltrationcurrentPre\right]$ is bounded due to the fact that $f \in C_{b}( \reals ^{d} \times \reals ^{d})$, then by dominated convergence theorem, we have
	
	\[
	\label{bounding term 2}
	\lim_{N \to \infty} \EE \left [\bigg| \frac{1}{N} \sum_{i=1}^{N} f\left(\CurrentSingleSample,  \CurrentSingleProposal\right)-\frac{1}{N} \sum_{i=1}^{N} \langle f, T \left(\CurrentSingleSamplePre,  \CurrentSingleProposalPre, \PreviousStepsize, \cdot\right) \rangle \bigg|\right ] = 0.
	\]
	Note that
	\[
	\begin{aligned}
		&\EE\left[\bigg| \frac{1}{N} \sum_{j=1}^{N} \langle f, T\left(\CurrentSingleSamplePre, \CurrentSingleProposalPre, \PreviousStepsize, \cdot, \cdot \right) \rangle-\frac{1}{N} \sum_{j=1}^{N} \langle f, T\left(\CurrentSingleSamplePre, \CurrentSingleProposalPre, \LimitingStepsizePre, \cdot, \cdot \right) \rangle \bigg|\right] \\
		& \leq \frac{1}{N} \sum_{j=1}^{N}\EE\left[\bigg|  \langle f, T\left(\CurrentSingleSamplePre, \CurrentSingleProposalPre, \PreviousStepsize, \cdot, \cdot \right) \rangle- \langle f, T\left(\CurrentSingleSamplePre, \CurrentSingleProposalPre, \LimitingStepsizePre, \cdot, \cdot \right) \rangle \bigg|\right] \\
		& = \EE\left[\bigg|  \langle f, T\left(X_{1}^{(t-1)}, Y_{1}^{(t-1)}, \PreviousStepsize, \cdot, \cdot \right) \rangle- \langle f, T\left(X_{1}^{(t-1)}, Y_{1}^{(t-1)}, \LimitingStepsizePre, \cdot, \cdot \right) \rangle \bigg|\right] .
	\end{aligned}
	\]
	Since we know $( X_{1}^{(t-1)}, Y_{1}^{(t-1)}, \PreviousStepsize) \convp (X_{1}^{(t-1)}, Y_{1}^{(t-1)},\LimitingStepsizePre)$ and $\langle f, T(x, y, h, \cdot, \cdot ) \rangle$ is bounded and continuous with respect to $(x,y,h )$, thanks to \cref{continuous and bounded mapping}, we have
	\[
	\label{bounding term 3}
	\lim_{N \to \infty} \EE\left[\bigg| \frac{1}{N} \sum_{j=1}^{N} \langle f, T\left(\CurrentSingleSamplePre, \CurrentSingleProposalPre, \PreviousStepsize, \cdot\right) \rangle-\frac{1}{N} \sum_{j=1}^{N} \langle f, T\left(\CurrentSingleSamplePre, \CurrentSingleProposalPre, \LimitingStepsizePre, \cdot\right) \rangle \bigg|\right] = 0 .
	\]

\subsubsection{Proof of \cref{lemma: second upper bounds}}

	Note that
	\[
	\begin{aligned}
		&\EE\left [\bigg| \frac{1}{N}\sum_{j=1}^{N}\langle f, T\left(\CurrentSingleSamplePre, \CurrentSingleProposalPre,\LimitingStepsizePre, \cdot, \cdot \right)\rangle-\left\langle f,\LimitingMeasureXY\right\rangle \bigg| ^{2}\right]\\
		& = \frac{1}{N} \EE\left [\bigg| \langle f, T\left(\CurrentSingleSamplePre, \CurrentSingleProposalPre, \LimitingStepsizePre, \cdot, \cdot \right)\rangle-\left\langle f,\LimitingMeasureXY\right\rangle \bigg| ^{2}\right]\\
		&+ \frac{N^{2}-N}{N^{2}} \EE \left[ \left( \langle f, T\left(X_{i}^{(t-1)}, Y_{i}^{(t-1)},\LimitingStepsizePre, \cdot, \cdot\right)\rangle-\left\langle f,\LimitingMeasureXY\right\rangle\right) \left( \langle f, T\left(\CurrentSingleSamplePre, \CurrentSingleProposalPre,\LimitingStepsizePre, \cdot, \cdot \right)\rangle-\left\langle f,\LimitingMeasureXY\right\rangle\right) \right].
	\end{aligned}
	\]
	The above expectation is taken with respect to $(\CurrentSingleSamplePre, Y_{i, t-1} )$. For any fixed $(\CurrentSingleSamplePre, Y_{i,t-1})$
	\[
	\begin{aligned}
		&\langle f, T\left(\CurrentSingleSamplePre, Y_{1, t-1}, \LimitingStepsizePre, \cdot\right)\rangle = \int f\left(s_{1}, s_{2} \right)  T\left(\CurrentSingleSamplePre, Y_{1, t-1}, \LimitingStepsizePre, ds_{1}, ds_{2}\right)\\
		&= \int f\left(\CurrentSingleSamplePre, s_{2} \right)\left(1-g\left( \CurrentSingleSamplePre, \CurrentSingleProposalPre, \LimitingStepsizePre\right)\right)Q_{ \LimitingStepsizePre}(\CurrentSingleSamplePre,  ds_{2})  \\
		& + \int f\left(\CurrentSingleProposalPre, s_{2} \right)g\left( \CurrentSingleSamplePre, \CurrentSingleProposalPre, \LimitingStepsizePre\right) Q_{ \LimitingStepsizePre}(\CurrentSingleProposalPre, ds_{2}).
	\end{aligned}
	\]
	By Assumption \ref{assumption1} and \ref{assumption2}, it is bounded and continuous with respect to $( \CurrentSingleSamplePre, \CurrentSingleProposalPre)$. Then thanks to that fact that $(X_{1:N}^{(t-1)}, Y_{1:N}^{(t-1)})$ is $\LimitingMeasureXYPre$-chaotic, we have
	\[
	\begin{aligned}
		& \lim_{N \to \infty}\EE \left[ \left( \langle f, T\left(X_{i}^{(t-1)}, Y_{i}^{(t-1)},\LimitingStepsizePre, \cdot\right)\rangle-\left\langle f,\LimitingMeasureXY\right\rangle\right) \left( \langle f, T\left(\CurrentSingleSamplePre, \CurrentSingleProposalPre,\LimitingStepsizePre, \cdot, \cdot \right)\rangle-\left\langle f,\LimitingMeasureXY\right\rangle\right) \right] \\
		& = \left( \int  \langle f, T\left(s_{1}, s_{2}, \LimitingStepsizePre, \cdot, \cdot \right)\rangle-\left\langle f,\LimitingMeasureXY\right\rangle  \dee \LimitingMeasureXYPre(ds_{1}, ds_{2}) \right)^{2}=0.
	\end{aligned}
	\]
	Since $f \in C_{b}( \reals ^{d} \times \reals ^{d} )$,  we have $\EE\left [\bigg| \langle f, T\left(\CurrentSingleSamplePre, \CurrentSingleProposalPre, \LimitingStepsizePre, \cdot, \cdot \right)\rangle-\left\langle f,\LimitingMeasureXY\right\rangle \bigg| ^{2}\right]\ < \infty$.
	Therefore, we have
	\[
	\label{bounding term 1}
	\lim_{N \to \infty} \EE\left[\bigg|\left\langle f, \EmpiricalMeasureXYTilde\right\rangle-\left\langle f,\LimitingMeasureXY\right\rangle \bigg| \right]
	= 0.
	\]

\subsection{Proof of \cref{prop:convergence of step size}}

	Recall that the step size update is 
	\[
	\begin{aligned}
		\CurrentStepsize &=\PreviousStepsize \exp \left( \frac{1}{t} \left(\left(\frac{1}{N} \sum_{j=1}^{N} \alpha_{j}^{(t-1)}\right)-\bar{\alpha}_{*}\right) \right) \\
		& = \PreviousStepsize \exp \left( \frac{1}{t} \left(\frac{1}{N} \sum_{j=1}^{N} g\left(\CurrentSingleSamplePre, \CurrentSingleProposalPre, \PreviousStepsize\right)-\bar{\alpha}_{*}\right) \right).
	\end{aligned}
	\]
	Note that
	\[
	\label{convergence of step size}
	\begin{aligned}
		&\frac{1}{N} \sum_{j=1}^{N} g(\CurrentSingleSamplePre, \CurrentSingleProposalPre, \PreviousStepsize) - \langle g(\cdot, \cdot,\LimitingStepsizePre), \LimitingMeasureXYPre \rangle\\
		&=\frac{1}{N} \sum_{j=1}^{N} g(\CurrentSingleSamplePre, \CurrentSingleProposalPre, \PreviousStepsize) - \frac{1}{N} \sum_{j=1}^{N} g(\CurrentSingleSamplePre, \CurrentSingleProposalPre, \LimitingStepsizePre) \\
		&\phantom{=~}+\frac{1}{N} \sum_{j=1}^{N} g(\CurrentSingleSamplePre, \CurrentSingleProposalPre, \LimitingStepsizePre) - \langle g(\cdot, \cdot, \LimitingStepsizePre), \LimitingMeasureXYPre \rangle.
	\end{aligned}
	\]
	The first term in \cref{convergence of step size} can be rewritten as
	\[
	\begin{aligned}
		&\EE \left[ \frac{1}{N} \sum_{j=1}^{N} g\left(\CurrentSingleSamplePre, \CurrentSingleProposalPre, \PreviousStepsize\right) - \frac{1}{N} \sum_{j=1}^{N} g(\CurrentSingleSamplePre ,\CurrentSingleProposalPre, \LimitingStepsizePre) \right] \\
		&= \EE \left[ g(\CurrentSingleSamplePre, \CurrentSingleProposalPre, \PreviousStepsize) -  g(\CurrentSingleSamplePre, \CurrentSingleProposalPre, \LimitingStepsizePre) \right].
	\end{aligned}
	\]
	Since $(\CurrentSingleSamplePre, \CurrentSingleProposalPre, \PreviousStepsize) \convp (\CurrentSingleSamplePre,\CurrentSingleProposalPre, \LimitingStepsizePre )$ and $g$ is bounded and continuous, it follows from \cref{continuous and bounded mapping} that 
	\[
	\label{limit of term 1}
	\lim_{N \to \infty}\EE \left[ \bigg| \frac{1}{N} \sum_{j=1}^{N} g(\CurrentSingleSamplePre, \CurrentSingleProposalPre, \PreviousStepsize) - \frac{1}{N} \sum_{j=1}^{N} g(\CurrentSingleSamplePre, \CurrentSingleProposalPre, \LimitingStepsizePre) \bigg| \right] = 0.
	\]
	The second term of \cref{convergence of step size} can be rewritten as 
	\[
	\frac{1}{N} \sum_{j=1}^{N} g(\CurrentSingleSamplePre, \CurrentSingleProposalPre, \LimitingStepsizePre) - \langle g(\cdot, \cdot, \LimitingStepsizePre), \LimitingMeasureXYPre \rangle\ = \langle g(\cdot, \cdot, \LimitingStepsizePre), \EmpiricalMeasureXYPre \rangle -\langle g(\cdot, \cdot, \LimitingStepsizePre),\LimitingMeasureXYPre \rangle.
	\]
	Since $(X_{1:N}^{(t-1)},  Y_{1:N}^{(t-1)})$ is $\LimitingMeasureXYPre$-chaotic, by \cref{propagation of chaos theorem} we have 
	\[
	\EmpiricalMeasureXYPre \convd  \LimitingMeasureXYPre.
	\]
	By Assumptions \ref{assumption1} and \ref{assumption2}, we have $g (x, y, \LimitingStepsizePre)$ is continuous with respect to $(x,y)$, thus 
	\[
	\label{limit of term 2}
	\langle g(\cdot, \cdot, \LimitingStepsizePre), \EmpiricalMeasureXYPre \rangle \xrightarrow{p} \langle g(\cdot, \cdot, \LimitingStepsizePre), \LimitingMeasureXYPre \rangle.
	\]
	It follows from \cref{limit of term 1} and \cref{limit of term 2} that 
	\[
	\frac{1}{N} \sum_{j=1}^{N} g(\CurrentSingleSamplePre, \CurrentSingleProposalPre, \PreviousStepsize) \xrightarrow{p} \langle g(\cdot, \cdot, \LimitingStepsizePre), \LimitingMeasureXYPre \rangle
	\]
	and therefore $\CurrentStepsize \convp \LimitingStepsize$. 

\subsection{Proofs of \cref{lemma:tightness lemma for X,lemma:subsequence distribution convergence lemma for X,lemma:chaotic for x}.}

The proof of \cref{lemma:tightness lemma for X} is similar to that of \cref{tightness lemma}; the proof of \cref{lemma:subsequence distribution convergence lemma for X} is similar to that of \cref{lemma:subsequence distribution convergence lemma}; the proof of \cref{lemma:chaotic for x} is similar to that of \cref{chaotic for the coupling}.

\section{ADDITIONAL RESULTS AND EXPERIMENTS}
\label{sec: ADDITIONAL RESULTS AND EXPERIMENTS}
\subsection{Ablation Study On $T$}
\label{sec: ablation study}
In \cref{sec:MC-size-parameters}, we discussed how to choose length of Markov chains properly: \cref{eq:chain-length} and \cref{eq:chain-length-HMC} guarantee that we can obtain efficient samples under different dimension $d$ without too large computation cost. Now, we would like to explore how the lower bounds would change under different number of iterations $T$. 
\begin{figure}
	\includegraphics[width=\linewidth]{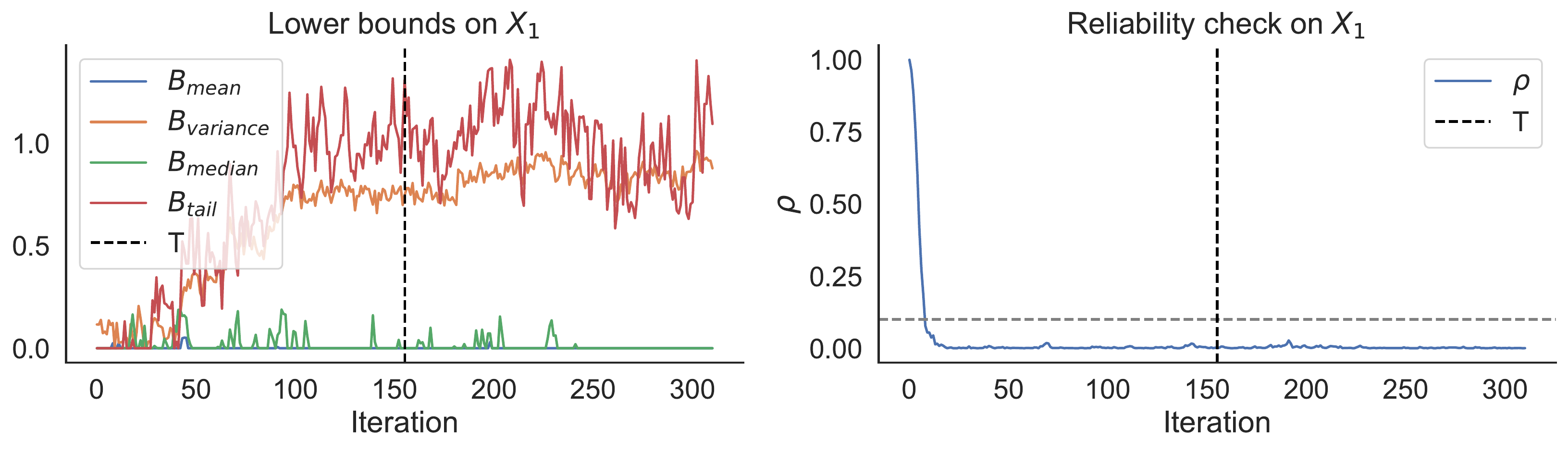}
	\caption{Ablation study for $T$ on $X_{1}$ of Gaussian model with correlated coordinates, $d=30$.}
	\label{fig: ablation study_gaussian}
\end{figure}

\begin{figure}
	\includegraphics[width=\linewidth]{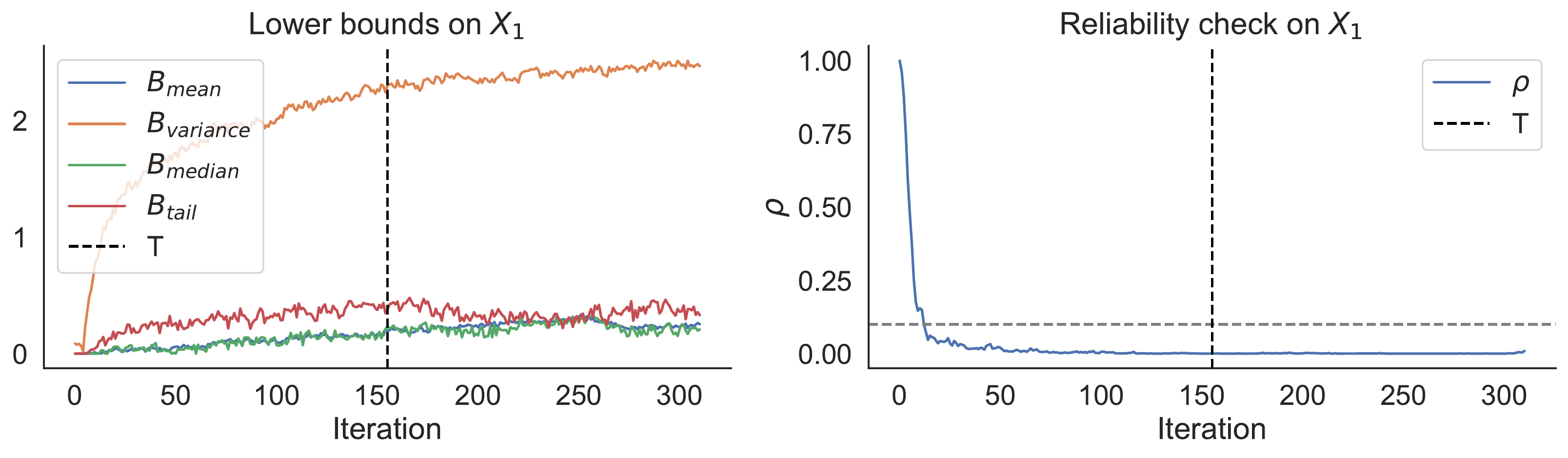}
	\caption{Ablation study for $T$ on $X_{1}$ of Neal-funnel shape model, $d=30$.}
	\label{fig: ablation study_neal}
\end{figure}

\begin{figure}
	\includegraphics[width=\linewidth]{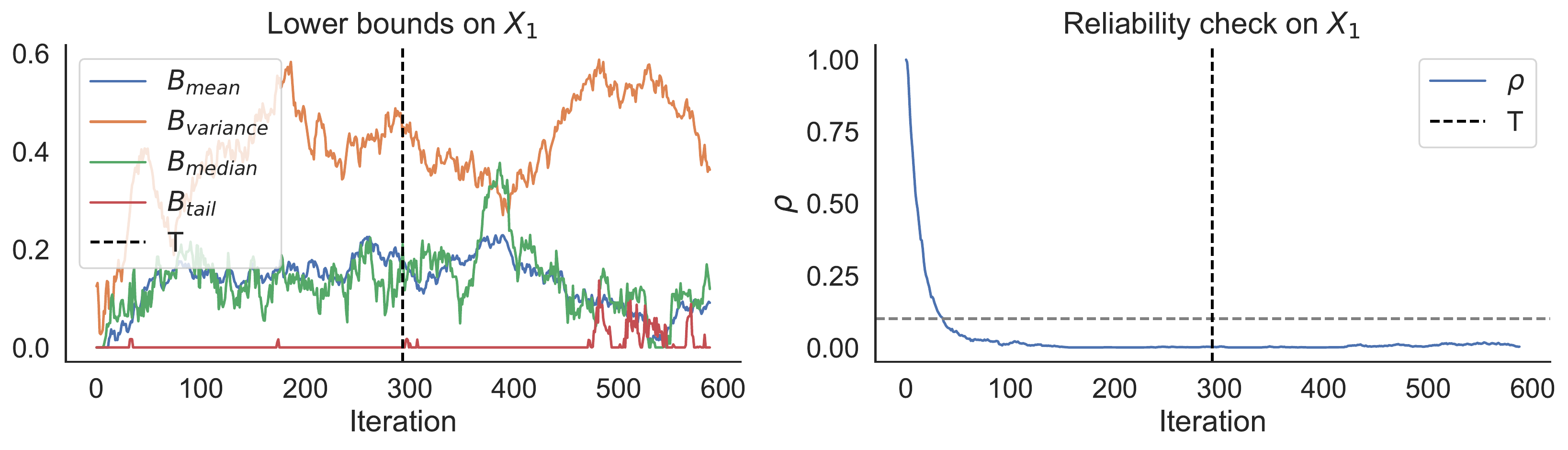}
	\caption{Ablation study for $T$ on $\lambda_{61}$ of Neal-funnel shape model, $d=100$.}
	\label{fig: ablation study_cancer}
\end{figure}

\Cref{fig: ablation study_gaussian,fig: ablation study_neal,fig: ablation study_cancer} display the evolution of several lower bounds for $X_{1}$ of Gaussian model (\cref{sec: Gaussian Model With Correlated Coordinates}), $X_{1}$ of Neal-funnel shape model (\cref{sec:Neal-Funnel Shape Model}) and $\lambda_{61}$ of cancer classification using horseshoe prior (\cref{sec: Cancer Classification Using a Horseshoe Prior}) respectively,
which shows that when $\rho$ is large (greater than 0.1), the proposed lower bounds are quite underestimated. 
These results also confirm the lowers bounds become nearly stationary at our proposed number of iterations $T$. 

\subsection{Additional Results for Neal-Funnel Shape Model}
\label{section: sm-neal}
\begin{figure}[t]
	\centering
	\includegraphics[width=.6\linewidth]{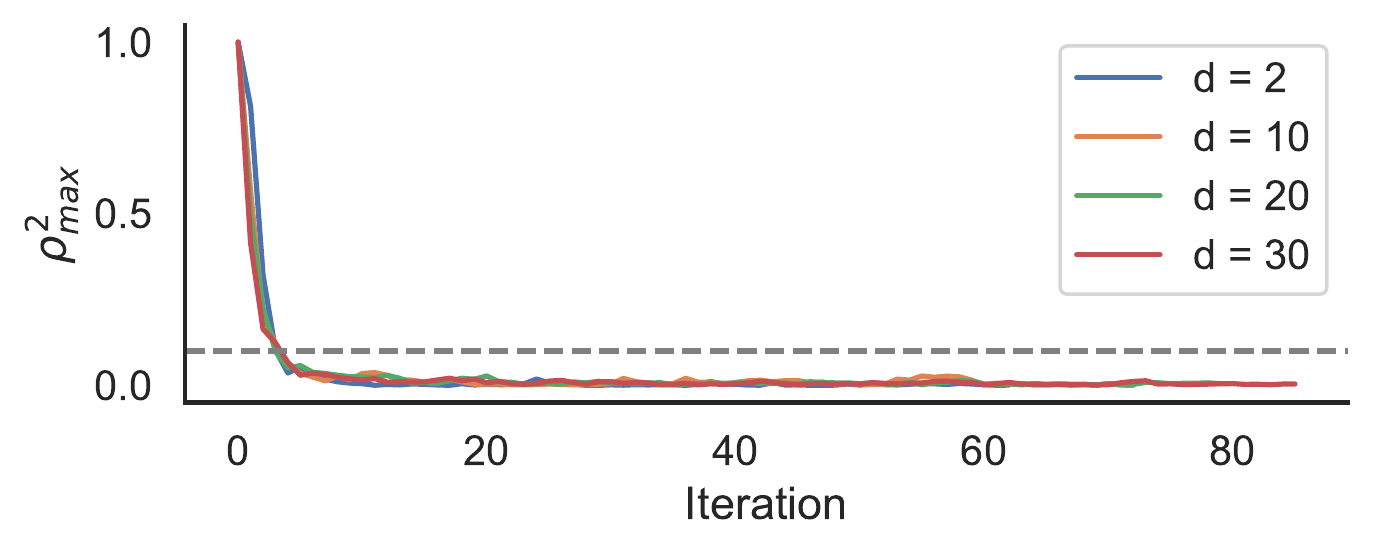}
	\caption{Reliability checks on TADDAA using Barker applied to the mean-field VI approximation to Neal's funnel distribution of varying dimensionality.}
	\label{fig:High Dimensional Neal Funnel reliability checks}
\end{figure}

In \cref{sec:Neal-Funnel Shape Model}, we validate \diagnosticName on Neal's funnel distribution approximated by mean-field Gaussian VI.
\cref{fig:High Dimensional Neal Funnel reliability checks} provides reliability checks for the Neal's funnel experiment \cref{fig:High Dimensional Neal Funnel Diagnostics}. 

Next, we will validate the \diagnosticName on Neal's funnel distribution approximated by mean-field t-distribution VI. The lower bounds $B_{\text{mean}}$ and $B_{\text{tail}}$ are computed according to \cref{eq:general lower bound} using functionals of interest $\mcF_{0.5}$ and $\mcF_{0.9}$ defined based on \cref{eq: quantile functionals of interest}.

\begin{figure}[t]
	\includegraphics[width=\linewidth]{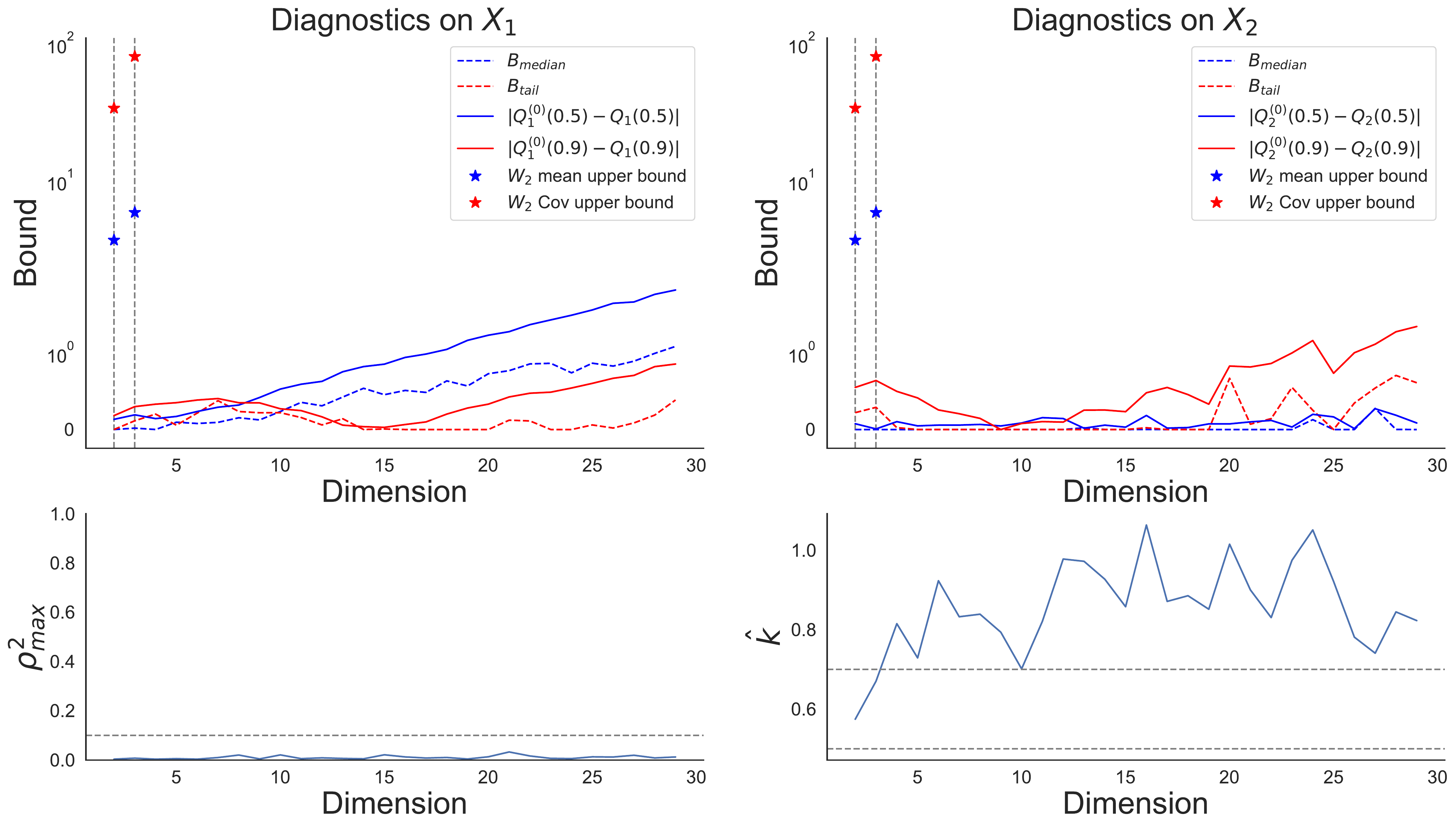}
	\caption{Diagnostics for Neal-funnel shape model, where 
		\diagnosticName uses the Barker proposal. 
		$Q_{1}(0.5)$ and $Q_{1}(0.9)$ denote the median and 90$\%$-quantile of $X_{1}$,  $Q_{2}(0.5)$ and $Q_{2}(0.9)$ denote the median and 90$\%$-quantile of $X_{2}$. }
	\label{fig:High Dimensional Neal Funnel Diagnostics-quantile}
\end{figure}
\cref{fig:High Dimensional Neal Funnel Diagnostics-quantile} shows that the median and tail quantile lower bounds constructed using Barker are quite precise and remain valid when $d$ and $hat{k}$ are large.	As we can see, $\hat{k}$ is large ($\hat{k}>0.7$) for almost any dimension, indicating VI approximation is not reliable. But if we are only interested in whether median of $X_{2}$, mean field t-distribution VI should be considered as reliable.
\subsection{Reliability Check for Bayesian Neural Network}
\label{sec: sm-bnn}

\begin{figure}[t]
	\centering
	\includegraphics[width=.6\linewidth]{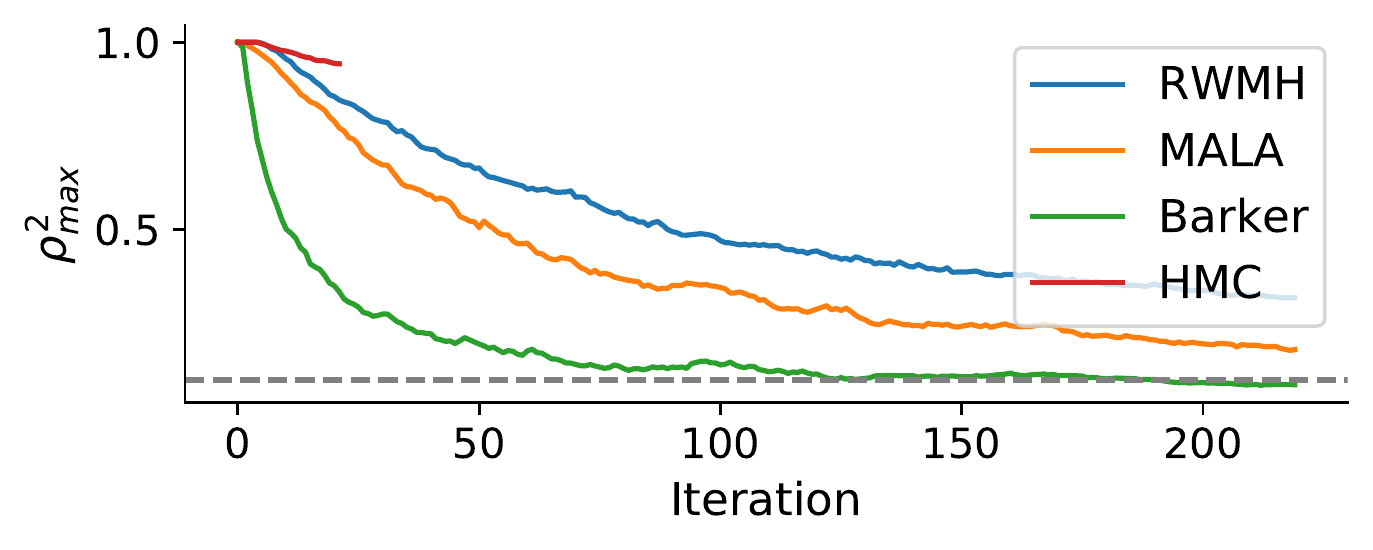}
	\caption{Reliability checks for \diagnosticName using different kernels applied to Bayesian neural network approximated by mean-field Gaussian VI for candy power ranking dataset.}
	\label{fig:Reliability checks for bnn}
\end{figure}

In \cref{sec: Bayesian Neural Network}, we validate \diagnosticName on Bayesian neural network approximated by mean-field Gaussian VI. 
The reliability check is shown in \cref{fig:Reliability checks for bnn}: 
due to the fact that the variance is dramatically underestimated, only Barker chains pass the reliability check, which is also consistent with the fact that Barker kernel provides the best lower bounds as displayed in \cref{fig:TADAA for candy classification_bnn.}.

\end{document}